# EPRA U-Net: An Efficient Pyramid Residual Attention Framework for Accurate Infarct Segmentation in Diffusion-Weighted MRI

Hasan Ulutas[1,*], Muhammet Emin Sahin[2,3], Mustafa Fatih Erkoc[4], Esra Yuce[1], Turker Tuncer[5], Sengul Dogan[5], Serkan Kiranyaz[6]

*[1]Yozgat Bozok University, Department of Computer Engineering, Yozgat, Turkey*
*[2]Izmir Bakircay University, Department of Computer Engineering, Izmir, Turkey*
*[3]Queen Mary's Digital Environment Research Institute (DERI), London, UK*
*[4]Yozgat Bozok University, Department of Radiology, Faculty of Medicine, Yozgat, Turkey*
*[5]Firat University, Department of Digital Forensics Engineering, Technology Faculty, Elazig, Turkey*
*[6]Qatar University, Department of Electrical Engineering, Doha, Qatar*
**Corresponding author: hasan.ulutas@yobu.edu.tr*

**Abstract— Objective:** Accurate identification of acute ischemic infarcts on diffusion-weighted magnetic resonance imaging (DWI) is a critical prerequisite for reliable lesion quantification and effective clinical decision support in the management of cerebrovascular events.

**Methods:** This study presents EPRA U-Net (Efficient Pyramid Residual Attention U-Net), a task-specific integrated architecture for efficient and accurate infarct segmentation of DWI images. In the proposed architecture, an EfficientNet-based encoder was used as a hierarchical feature extractor with a minimized parameterization. In addition, a Residual-Recurrent (R2) block (recurrent unrolling step $t = 2$, following the original formulation) and Atrous Spatial Pyramid Pooling (ASPP) were integrated to enhance the performance of spatial dependency modeling. Additionally, a dual attention mechanism was incorporated to highlight lesion-related activations while concurrently enabling the suppression of extraneous background responses. To prioritize lesion detection consistent with clinical imperative, a Tversky loss function ($\alpha = 0.4$, $\beta = 0.6$) was adopted, thereby emphasizing the sensitivity of detection over its specificity during the optimization process.

**Results:** Experimental evaluations were conducted utilizing an in-house dataset comprising 167 patients with 4,895 DWI slices; subsequently, the performance of the proposed EPRA U-Net was assessed in comparison with state-of-the-art models, specifically UNet++, DeepLabV3+, and TransUNet. The experimental results suggest that EPRA U-Net attained superior performance, evidenced by a pixel-aggregated Dice of 0.8984, a per-sample Dice of 0.9469, an IoU of 0.8155, a Recall of 0.8887, a Lesion F1 of 0.9378, and an HD95 of 11.62 px (median: 2.00 px). Furthermore, a clear reduction in the rate of missed lesions, specifically by 16%, 25%, and 29%, was observed when compared with UNet++, DeepLabV3+, and TransUNet, respectively. Additionally, a comprehensive ablation study was systematically conducted.

**Conclusion:** These findings suggest that the proposed architecture affords a more robust and accurate infarct segmentation, consequently facilitating a more reliable estimation of treatment eligibility.

## 1. Introduction

Ischemic strokes, also referred to as cerebral infarctions, result from the obstruction of cerebral blood flow by vascular blockage, culminating in tissue hypoxia and cell necrosis [1]. According to the World Stroke Organization (WSO), cerebrovascular events constitute the second leading cause of global mortality and the third leading cause of death and disability combined [2]. Early detection is instrumental in facilitating accurate stroke diagnosis and the prompt implementation of therapeutic interventions [3]. Infarcts detected within six hours are designated acute infarcts; subacute infarcts are discerned 24 to 48 hours post-onset; those beyond this timeframe are classified as chronic infarcts. This temporal categorization, however, significantly influences both stroke treatment protocols and patient prognoses. Diffusion-weighted magnetic resonance imaging (DWI) has emerged as a gold-standard method for early detection of ischemic infarcts, as this modality permits detection of ischemia-induced alterations in the diffusion of water molecules within minutes following infarct onset [4]. The manual identification of infarct areas, based on DWI, is a time-intensive process, with potential for substantial inter-reader variability of infarct delineation. Consequently, such inherent challenges have stimulated interest in the recent development of computer-assisted stroke detection tools [5]. This investigation aims to devise a highly accurate deep learning (DL)-based tool for infarct area segmentation in DWI images, which may serve as a valuable adjunct to clinician decision-making during stroke assessment. To this end, a dataset of infarcts was curated at this research institution; an original model architecture (EPRA U-Net) was proposed and rigorously evaluated against state-of-the-art approaches; and an evaluation methodology, appropriate for deployment in the clinical environment, was devised.

In recent years, deep learning algorithms have demonstrated substantial advancements in medical image segmentation. A landmark study by Ronneberger et al. [6] introduced a U-Net architecture, which combined an encoder-decoder structure with skip connections, facilitating effective model training even on relatively modest datasets. Subsequent advancements within this domain aimed to improve feature representation and reduce the semantic gap between input and target layers. Specifically, Nested UNet++ [7] incorporated nested skip connections, whereas Attention U-Net [8] incorporated attention gates to emphasize pertinent anatomical structures. R2U-Net [9] then leveraged residual and recurrent components to reinforce feature propagation.

Transformers also enhanced the performance of models in segmentation tasks. TransUNet [10] used a combination of Vision Transformers with CNNs to introduce context to models, while hierarchical transformers were incorporated into Swin-UNet architecture for multi-scale feature representation [11]. Despite these advantages of transformer architectures, notably, transformers may exhibit computational intensity, and overfitting could manifest on relatively small datasets [12]. Furthermore, 3D models have been proposed to provide additional contextual information through the comprehensive processing of volumetric data. 3D U-Net [13] added a 3D component to U-Net architecture, while V-Net [14] employed dice-based optimization for mitigating class imbalance issues. Multi-scale 3D convolutional neural network architectures, such as DeepMedic [15], showed high accuracy in segmenting brain lesions. However, using 3D modeling may significantly increase computational expenditure, given the need to process each voxel across all spatial dimensions [16]. Another U-Net model, nnU-Net [16], functioning as a novel framework, addressed the need for automated model development by formulating an algorithm capable of adapting segmentation pipelines according to the target dataset. Contemporary advancements have primarily focused on multi-scale modeling to advance segmentation methodologies. In DeepLabv3+, Atrous Spatial Pyramid Pooling (ASPP) provided receptive field extension using dilated convolutions, which aided in improving boundary delineation [17]; concurrently, PSPNet [18] aggregated global information via pyramid pooling structure. Attention mechanisms have emerged as a fundamental component in segmentation models following their initial introduction within transformer architectures [19]. Attention-Based Methods for Image Segmentation, such as CBAM [20], sequentially employ channel and spatial attention, and compute dual (position and channel) attention simultaneously, as in DANet [21]. A primary impediment to the successful segmentation of infarcts is the significant class imbalance inherent in these datasets. Dice objectives may alleviate certain class balance issues; conversely, Generalized Dice Loss [22] was developed to enhance model stability when confronted with imbalanced data [23]. The Tversky loss function [24] introduced $\alpha$ and $\beta$ weights, thereby regulating the equilibrium between the costs associated with false positives and false negatives; notably, higher $\beta$ values tend to favour the reduction of false negatives—an especially advantageous characteristic for clinically important diagnostic decisions such as stroke detection [25]. Focal Tversky Loss [26] has been demonstrated to potentially improve performance when applied to challenging and small lesions. In summary, this study proposes the development of an optimized infarct segmentation model, which is primarily driven by the need for the clinical application of automated stroke imaging solutions and the identifiable deficiencies inherent in existing segmentation methodologies. Instead of relying on architectural innovation alone, the proposed approach combines several complementary approaches such as efficient encoding, multi-scale modeling, recurrent modeling, and attention mechanisms in a framework specifically tailored to overcome infarct segmentation challenges. Furthermore, a significant strength of this study lies in generating a singular, clinically representative dataset, alongside using multi-layered statistical validation for the rigorous assessment of model performance. Despite numerous advancements in brain lesion segmentation during recent years, including establishing a BraTS benchmark for tumor segmentation [27], comparatively limited research has been conducted concerning DWI infarct segmentation [28].

## 1.1. Motivation and Contributions

Critical analysis of existing approaches for solving the problem at hand suggests the presence of two significant deficiencies in this body of research: reliance on innovative architecture without sufficient statistical validation[29] and neglecting clinically motivated segmentation loss functions that address class imbalances and asymmetry in costs of incorrect predictions[30]. In light of the above considerations, novel and significant contributions of this paper can be summarized as follows:

1. An original infarct dataset was developed based on data from Yozgat Bozok University Hospital, including 167 patients with 4,895 annotated slices under clinically realistic class-imbalance conditions (14.7% positive slice rate).
2. A novel and potentially state-of-the-art segmentation model was introduced, the EPRA U-Net, which represents a principled hybrid architecture that integrates EfficientNet-based hierarchical encoding, recurrent-residual spatial modeling, ASPP-based multi-scale context modeling, and a dual attention mechanism.
3. We applied a clinically motivated Tversky loss function configuration ($\beta > \alpha$) to address the asymmetrical costs of incorrect predictions for infarct segmentation;
4. Finally, we created a complete validation scheme, including non-parametric paired testing, pixel-level McNemar analysis, bootstrap confidence interval computation, and effect size estimation.

As such, the contributions collectively aim to advance infarct detection technology by providing a comprehensive framework that incorporates both architectural innovation and clinically relevant segmentation objectives, concurrently employing rigorous statistical analysis techniques for evaluation. The remaining sections outline the organisational structure: Section 2 presents data preparation, model architecture, evaluation metrics, and the statistical framework. Section 3 provides a comprehensive account of experimental outcomes, encompassing quantitative assessments, comparative evaluations including error analysis, distributional analysis, and statistical analysis results. Section 4 discusses the findings in relation to existing literature, analyses the architectural selection, and addresses the limitations inherent in this work. Section 5 outlines the concluding remarks and prospective directions.

## 2. Material and Methods

### 2.1. Dataset

The dataset employed in this investigation was retrospectively acquired from the Department of Radiology at Yozgat Bozok University Training and Research Hospital. Before the analysis stage, compliance with institutional ethical standards was ensured, and all patient data were fully anonymized. Only Diffusion-Weighted Imaging sequences were incorporated into the study, as DWI is considered the established standard for the detection of acute ischemic stroke lesions, owing to its superior sensitivity to restricted water diffusion within infarcted tissue [31]. Next, a comprehensive analysis encompassed 4,895 slices, which were derived from the DWI sequences of 167 patients exhibiting a diagnosis of acute ischemic stroke. Image annotation was performed using the MakeSense platform (makesense.ai) [32], which is a web-based tool that allows precise polygon-level annotation. A radiologist specializing in neuroimaging annotated all images; furthermore, to enhance annotation reliability, researchers applied a two-stage consistency check, independently annotating each image across two distinct sessions. Consequently, only those slices demonstrating consistent annotations in both sessions were incorporated into the final dataset, mitigating the risk of labeling errors attributable to lesion ambiguity and reviewer fatigue. The dataset also exhibits a class imbalance. Among all slices, 719 (14.7%) contained infarct lesions, whereas 4,176 (85.3%) were normal. The median lesion size was 221 pixels, with an interquartile range of 90–699 pixels. These findings may suggest that the majority of infarcts were relatively small and exhibited complex morphologies. Fig. 1 presents the dataset division scheme.

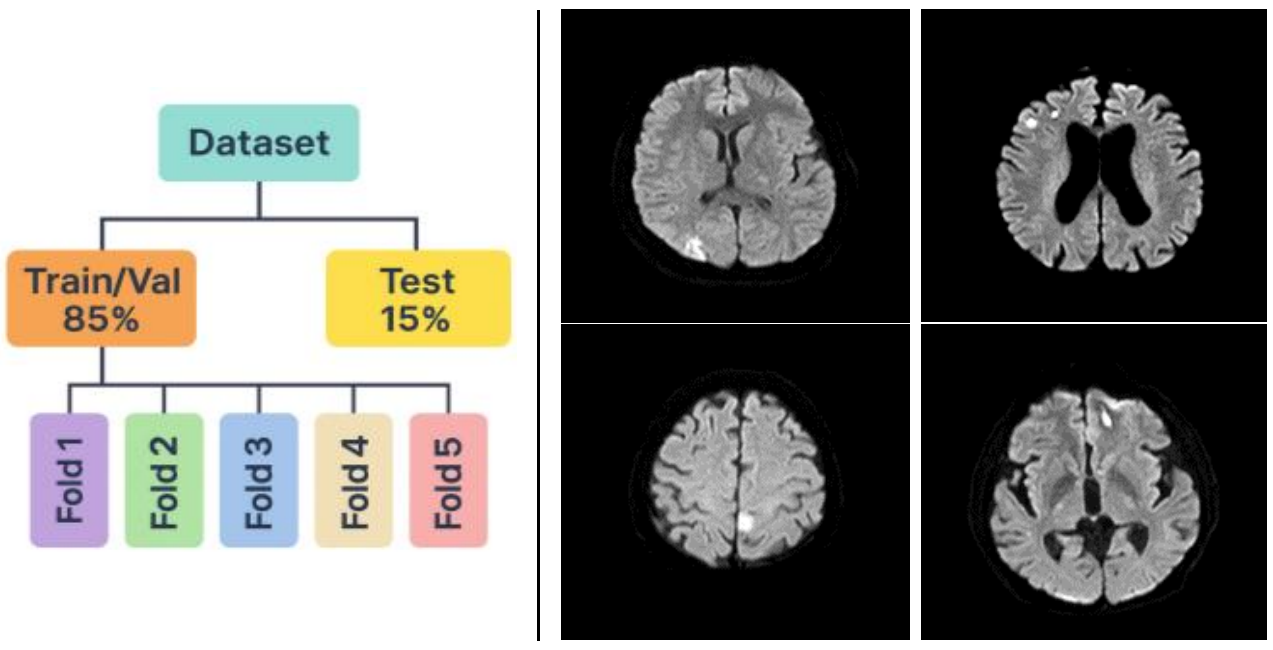

Fig. 1. The patient-level dataset partitioning strategy is clarified; furthermore, representative Diffusion-Weighted Imaging (DWI) slice examples are presented, each accompanied by its corresponding ground-truth data.

The prevention of potential data leakage and the assurance of unbiased model evaluation required splitting data at the patient level rather than the slice level. This distinction is critical in biomedical image analysis because slice-based splitting may place related slices from the same patient into both the training and test sets. Such a case may lead to an artificial increase in performance because of the similarity in anatomy and scanning conditions [33].

The implementation of patient-level splitting ensured that all slices originating from a single patient were allocated exclusively to one partition. This approach established the mutual exclusivity of the training, validation, and test sets. A stratified sampling strategy was implemented to achieve an approximate 70%, 15%, and 15% distribution for the training, validation, and test sets, respectively. Accordingly, 116 patients were assigned to the training set, 25 patients to the validation set, and 26 patients to the test set. Furthermore, stratified sampling was employed to maintain, as far as practicable, the original lesion-to-normal distribution. This approach, therefore, facilitated the preservation of the significant class imbalance inherent in the dataset. Consequently, none of the subsets exhibited unrealistic overrepresentation of one class, and the overall data distribution appeared to align more closely with authentic clinical conditions. The dataset used in this study was retrospectively collected from the Department of Radiology at Yozgat Bozok University Training and Research Hospital. This study was approved by the Clinical Research Ethics Committee of Yozgat Bozok University (Protocol No:36/27). Due to the retrospective nature of the study and the use of fully anonymized data, informed consent was waived.

## 2.2. Data Pre-processing and Augmentation

The pre-processing pipeline comprised four distinct stages: format conversion (incorporating intensity windowing); 2.5D slice stacking; intensity normalisation; and training-time data augmentation. The pre-processing and augmentation pipeline is shown in Fig. 2.

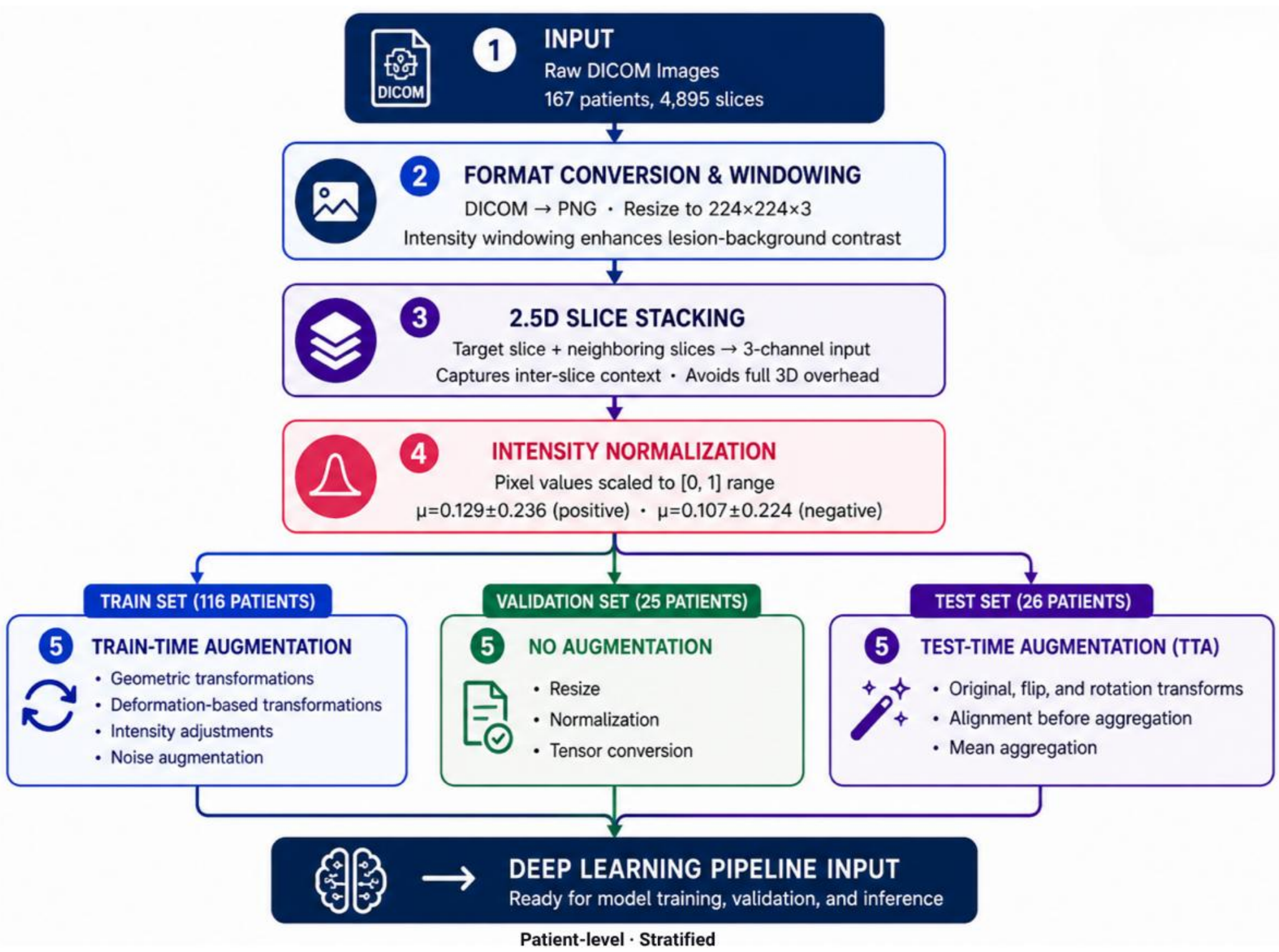

Fig. 2. Data pre-processing and augmentation pipeline.

**Image Format** Conversion and Windowing: The original Digital Imaging and Communications in Medicine (DICOM) images underwent conversion into 224×224×3 Portable Network Graphics (PNG) images. Intensity windowing simultaneously applied during this phase. This procedure augmented the contrast between lesion and background regions through the exclusion of superfluous intensity ranges. Ischemic stroke lesions frequently manifest as hyperintense regions in diffusion-weighted images, thus warranting distinct separation from normal brain tissue [4].

**2.5D Method:** Rather than employing individual two-dimensional (2D) slices, a 2.5D methodology was adopted. In this method, each target slice was combined with its neighboring slices to form a 3-channel input. This strategy facilitated the representation of inter-slice contextual information whilst maintaining the computational parsimony inherent to 2D convolution. Compared with three-dimensional (3D) models, its implementation required less memory and a reduced computational expenditure [34]. Given the considerable scale of this dataset, the 2.5D approach may therefore represent a particularly effective choice.

**Normalization:** Subsequent to the preparatory phase of model training, all input images were subjected to intensity normalization, a process undertaken ensuring numerical stability and a consistent distribution of inputs across the dataset. Specifically, pixel intensities were linearly scaled to the [0, 1] range according to their bit depth: eight-bit images (uint8) were consequently divided by 255, whilst sixteen-bit images (uint16) were divided by 65535. This fixed-range min-max normalization can be formally expressed as: $x$_norm =$x$/$x$_max, where $x_{max} \in \{255, 65535\}$ depending on the acquisition bit depth. Notably, this methodology entails the application of a globally fixed scaling factor, in contrast to per-image adaptive normalization (e.g., min–max rescaling based on each image's intrinsic intensity range), ensuring the preservation of inter-patient and inter-session intensity relationships in the original dynamic range characteristic of the imaging modality.

Although the EfficientNet-B0 encoder was initialized with ImageNet pre-trained weights, the conventionally recommended channel-wise ImageNet normalization ($\mu$ = [0.485, 0.456, 0.406], $\sigma$ = [0.229, 0.224, 0.225]) was deliberately not applied. This decision is grounded in the 2.5D input design: the three input channels correspond to consecutive DWI slices (previous, current, and next), rather than RGB colour channels. Applying RGB-derived population statistics to single-modality grayscale DWI slices would introduce a systematic intensity bias incompatible with the imaging modality. The adopted [0, 1] linear scaling is a modality-appropriate preprocessing choice and permits the encoder's first-layer representations to adapt to DWI-specific intensity distributions during fine-tuning, consistent with established practice in MRI segmentation [16]. An analysis of intensity distribution across the dataset indicated a mean intensity of 0.129 ± 0.236 for positive (infarct-containing) slices, contrasting with 0.107 ± 0.224 for negative (normal) slices. The higher mean intensity in positive slices is consistent with the characteristic hyperintense appearance of acute ischemic lesions on DWI, which stems from restricted water diffusion in infarcted tissue [4]. The relatively substantial standard deviation values demonstrated across both cohorts largely reflect the inherent heterogeneity related to lesion extent, slice positioning, and patient-specific imaging parameters. Despite the modest absolute disparity between the two cohorts, this intensity discrepancy may contribute a subtle yet non-trivial discriminative signal that might complement the learned spatial features extracted by the segmentation models.

**Data Augmentation:** To enhance model generalization and mitigate overfitting, several augmentation methods were applied exclusively to the training data; these encompassed flipping, rotation (±15°), scaling (0.9 to 1.1), and elastic deformation. The initial three methods generated geometric variations, which serve to reflect natural inter-patient differences, whereas elastic deformation facilitated the modeling of infarct lesion shape variability.

## 2.3. Motivation for the Proposed Architecture

Acute infarct segmentation on diffusion-weighted imaging, an inherently difficult task, presents several specific challenges, suggesting that a general segmentation architecture may not suffice. Encoder-decoder models such as UNet and its variants have shown strong performance in many biomedical image segmentation tasks. Nonetheless, certain limitations are evident in DWI-based acute infarct segmentation, which may warrant specific architectural considerations. In UNet, skip connections transfer spatial information, but low-level noise is also transferred. The issue is exacerbated when lesion boundaries lack clear delineation and image signals exhibit heterogeneity. UNet++ partially addresses this issue with its nested skip pathways. Nonetheless, it does not directly facilitate the expansion of multi-scale receptive fields. Transformer-based models such as TransUNet can learn long-range dependencies effectively. However, their implementation typically incurs a higher computational cost, and they may exhibit a propensity for overfitting, particularly on small, single-center datasets, such as the one employed in this investigation.

## 2.4. Proposed Model: EPRA U-Net

The EPRA U-Net is a mixed deep learning architecture specifically engineered for brain infarct segmentation; its functional composition is shown in Fig. 3. Four key components underpin the architecture, addressing the challenges associated with the image recognition of ischemic strokes:

**EfficientNet-based Encoder:** For its encoder component, the model employs EfficientNet-B0, developed by Tan and Le [35]. In comparison with ResNet-50, EfficientNet-B0 exhibits an approximate 79.3% reduction in parameters, yet it may attain approximately 7% higher performance through using a compound scaling strategy that effectively balances width, depth, and resolution. Notably, EfficientNet is based on MBConv and Squeeze-and-Excitation modules, which facilitate the efficient extraction of features.

**Residual-Recurrent (R2) Blocks:** The model incorporates residual connections, proposed by He et al. [36], alongside recurrent convolutional connections from the R2U-Net architecture [9]. This structural integration facilitates enhanced information flow throughout the model, which may contribute to the more effective segmentation of complex and small lesions.

**Atrous Spatial Pyramid Pooling (ASPP):** The Atrous Spatial Pyramid Pooling approach, as delineated by Chen et al. [37], enables the model to use atrous convolutions with various dilations. This mechanism enables the expansion of the receptive field, thereby acquiring both local and global feature representations concurrently.

**Dual Attention Mechanism:** The EPRA U-Net incorporates a parallel position-channel dual attention mechanism, derived from Transformer self-attention principles, which facilitates the modeling of pixel and feature map dependencies. This, in turn, may enable an automated focus on lesion areas in the image, thereby potentially enhancing the quality of segmentation.

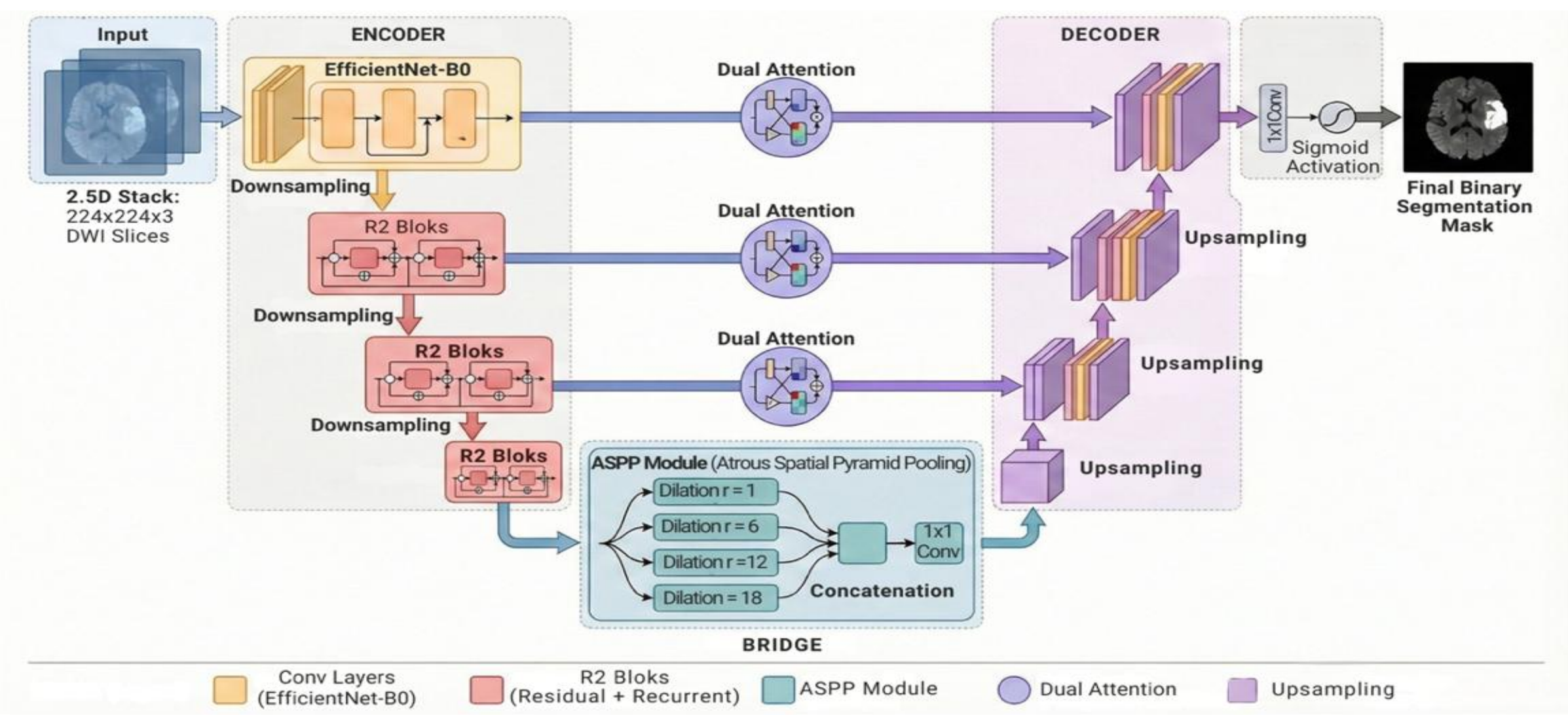

Fig. 3. Detailed functional components of EPRA U-Net.

As illustrated in Fig. 3, the architectural pipeline of the proposed EPRA U-Net model comprises the following stages:

***Step 1:*** *The initial stage involves preparing the input as a 2.5D stack (224 × 224 × 3) by combining the target Diffusion-Weighted Imaging slice with its adjacent slices.*

***Step 2:*** *Next, the input is transmitted to the encoder to extract first-level features using the EfficientNet-B0 backbone.*

***Step 3:*** *Downsampling is then performed to reduce spatial size and generate more compact feature maps.*

***Step 4:*** *The downsampled features are then transmitted through the initial R2 block, which refines spatial representations using residual and recurrent operations.*

***Step 5:*** *The downsampling and R2 block operations are reiterated at progressively deeper levels to acquire more robust and abstract lesion-related features.*

***Step 6:*** *The deepest encoder features are transferred to the network's designated bridge component.*

***Step 7:*** *Within the bridge, the ASPP module extracts multiscale contextual information using parallel dilated convolutions, each with distinct dilation rates.*

***Step 8:*** *Following this, the ASPP outputs are concatenated, and are then fused via a 1 x 1 convolutional operation.*

***Step 9:*** *The encoder features are then transferred to the skip pathways to be then processed by dual attention modules.*

***Step 10:*** *This dual attention mechanism may augment lesion-related responses while simultaneously mitigating irrelevant background activations.*

***Step 11:*** *The decoder's operation begins with the ASPP output, where the feature maps are sequentially upsampled.*

***Step 12:*** *Each upsampled decoder feature map is then fused with its corresponding attention-refined skip feature from the encoder, thus integrating refined information.*

***Step 13:*** *The upsampling process continues until the feature maps reach the required output resolution.*

***Step 14:*** *Next, a 1 x 1 convolution is applied, to convert the final decoder features into a single-channel segmentation map.*

***Step 15:*** *Sigmoid activation converts the output values into pixel-wise probabilities.*

***Step 16:*** *Ultimately, the final binary segmentation mask is generated, distinguishing the infarct region from the surrounding background.*

### 2.4.1. Mathematical Modelling: Foundational Formulations

This subsection presents the foundational mathematical formulations underpinning the pivotal components of the EPRA U-Net architecture.

**EfficientNet Compound Scaling:** The EfficientNet-B0 encoder employs a strategy of compound scaling for the uniform scaling of depth, width, and resolution through using a singular coefficient, φ. The scaling factors are defined as follows: $d = \alpha\phi, w = \beta\phi, r = \gamma\phi$, subject to the constraint $\alpha \cdot \beta^2 \cdot \gamma^2 \approx 2$, where d, w, and r denote the network depth, width, and resolution multipliers, respectively, and $\alpha_E = 1.2, \beta = 1.1, \gamma = 1.15$ for EfficientNet-B0 [35].

**Residual-Recurrent (R2) Block:** In each R2 block, the recurrent convolution operation at time step t is described by the output $z_{ijk}(t) = f((w_f^k)^T x_f^{(i,j)}(t) + (w_r^k)^T x_r^{(i,j)}(t-1) + b_k)$, where $x_f^{(i,j)}(t)$ and $x_r^{(i,j)}(t-1)$ are the feedforward and recurrent inputs respectively, $w_f^k$ and $w_r^k$ are their corresponding weights, $b_k$ is the bias, and $f(\cdot)$ denotes the Rectified Linear Unit (ReLU) activation function [9]. Next, the residual connection is applied, integrating the input to the block with the output generated from the recurrent layers [38].

**Atrous Spatial Pyramid Pooling (ASPP):** The Atrous Spatial Pyramid Pooling module applies parallel atrous (dilated) convolutions employing varied dilation rates (1, 6, 12, and 18) for the capture of multi-scale contextual information[37]. Formally, the output of an atrous convolution, characterized by a specific dilation rate d, is computed as follows: $f(x,y) = \sum_k \sum_l \sum_m I\,(x + rk, y + rl, m) W(k,l,m) + B(m)$, where I represents the input image, (x, y) denotes the spatial position, W and B constitute the weights of the kernel, and r specifies the atrous rate [37]. Next, the outputs emanating from all branches—encompassing a one by one convolution and global average pooling—undergo concatenation and following merging with an additional convolution operation, which yields a compact multi-scale feature map [37].

**Dual Attention Mechanism:** The dual attention module operates along two parallel branches: position attention and channel attention. In the position attention branch, given an input feature map $A \in R^{C \times H \times W}$, the spatial attention map $S \in R^{N \times N}$ (where N = H × W) is computed. The output of the position attention is: $E_j = \alpha \sum_i (s_{ij} \cdot D_i) + A_j$, where α is a learnable scaling parameter initialized to 0. Similarly, the channel attention branch computes inter-channel dependencies: $x_{ij} = exp(A_i \cdot A_j)/\Sigma_k exp(A_i \cdot A_k)$ and the channel-attended output is: $E_j^c = \beta \sum_i (x_{ij} \cdot A_i) + A_j$. The aggregation of the dual attention mechanism's output appears to be the element-wise summation of both these two branches [21].

**Squeeze-and-Excitation (SE) Module:** Within the architectural configuration of EfficientNet MBConv blocks, the Squeeze-and-Excitation module performs the recalibration of channel-wise feature responses [35]. Given an input feature map $U \in R^{C \times H \times W}$, global average pooling is first applied to produce a channel descriptor $z \in R^C$, where $z^c = \left(\frac{1}{H} \times W\right) \sum_i \sum_j u^c(i,j)$. The excitation output is computed as: $s = \sigma(W^2 \cdot \delta(W^1 \cdot z))$, where $W_1 \in R^{\frac{C}{r} \times C}$ and $W_2 \in R^{C \times \frac{C}{r}}$ are fully connected layers with reduction ratio r, δ is the ReLU activation, and σ is the sigmoid function. The final output is the channel-wise multiplication: $\hat{x}^c = s^c \cdot u^c$.

### 2.4.2. Training Strategy and Loss Function

The optimization of the Efficient Pyramid Residual Attention U-Net (EPRA U-Net) is primarily driven by the Tversky Loss Function [24], which generalizes the Dice score, allowing for refined control over both false positives (FP) and false negatives (FN), as described below:

$$TI = \frac{TP}{TP + \alpha \times FP + \beta \times FN} \tag{1}$$

Reflecting the clinical asymmetry between false negatives and false positives at both the optimization and evaluation levels, adopting the Tversky loss (α = 0.4, β = 0.6) as the training objective and an additional assessment of model performance was conducted utilising the F2 score, which assigns greater weight to recall than precision, reflecting the preference for high recall to minimise missed detections [25]. This dual emphasis is intended to ensure that the model's tendency to minimising missed detections is consistently enforced during training and transparently quantified during evaluation. Training hyperparameters are given in Table 1.

$$F2 = \frac{5 \times TP}{5 \times TP + FP + 4 \times FN} \tag{2}$$

Table 1. Training hyperparameters

| Parameter | All Models (except of nnUnet) |
|---|---|
| Number of Epochs | Varies by model |
| Batch Size | 16 |
| Optimizer | AdamW |
| Initial Learning Rate | 1e-4 |
| Weight Decay | 1e-5 |
| Learning Rate Scheduler | CosineAnnealing |
| Early Stopping | Patience = 15 |

The training and validation losses were monitored across all 60 epochs for the evaluation of convergence behavior and the mitigation of overfitting phenomena. EPRA U-Net was allocated 60 training epochs to accommodate the increased convergence time imposed by its composite architecture, comprising an EfficientNet-B0 encoder, ASPP, R2 blocks, and dual attention mechanism. Baseline models converged within 50 epochs, as confirmed by the early stopping criterion (patience = 15), which was applied uniformly across all models; the epoch ceiling difference therefore reflects architectural complexity rather than a systematic training advantage. The model checkpoint with the best validation Dice score was chosen for further testing on the separate test dataset. The Dice score was used as the primary criterion for selecting the final model because it directly corresponds to the volumetric overlap — a critical objective in segmentation of lesions [39] — and is insensitive to unbalanced classes, which may represent an inherent characteristic of stroke datasets where lesions are only a minimal proportion of the overall volume [40]. Besides the Dice score, other measures, such as F2 score, and lesion-based F1, were presented; however, the model checkpoint selection was based on the former because it is a preferred metric for assessing segmentation performance [40].

### 2.4.3. Comparative Benchmark Models

To evaluate the performance of the proposed EPRA U-Net, a comparative analysis was conducted employing three state-of-the-art segmentation models:

- UNet++ [7]: Incorporates nested skip connections to reduce the semantic gap between the encoder and decoder
- TransUNet [10]: Consists of both Vision Transformer (ViT) and CNN architectures for utilizing high-dimensional information from the global context.
- DeepLabV3+ [17]: Implements ASPP layers inside the encoder-decoder framework to improve contour recognition

**nnU-Net (2D) [16]:** A self-configuring framework that automatically adapts its preprocessing, network architecture, and training pipeline according to dataset characteristics, without requiring manual hyperparameter tuning. For this study, the 2D nnU-Net configuration was employed using the same 3-channel 2.5D input representation. nnU-Net was trained under its default protocol (1,000 epochs; batch size 65; Dice and Cross-Entropy compound loss; SGD with Nesterov momentum 0.99; polynomial LR decay; automatic ZScore normalisation) on the same training set and evaluated on the identical held-out test set.

## 2.5. Evaluation Metrics

Evaluation of segmentation results is imperative for medical image analysis because any errors in the segmentation process may exert direct consequences on clinical decisions [41]. Prior scholarly investigations indicate that no singular evaluation metric is capable of encompassing the entirety of segmentation quality's characteristics, especially in instances involving class imbalance and anomalous object geometries, which are common in biomedical images [42]. Consequently, given this observation, this study employs multiple measures that focus on region overlap, contour precision, confidence of classification, and pixel-wise prediction accuracy [43],[44],[45].

These measures are described below: The Dice Similarity Coefficient (DSC) provides an estimation of the similarity between the predicted segmentation mask and the ground-truth image and is widely employed in medical image segmentation owing to its insensitivity to class imbalance [46]. DSC is calculated as follows:

$$\text{Dice} = \frac{2\mid P \cap G \mid}{\mid P \mid + \mid G \mid} \tag{3}$$

Herein, P denotes the predicted mask, whereas G represents the ground truth. Elevated Dice scores typically suggest a stronger agreement between the predicted masks and ground truths. The Intersection over Union (IoU), also known as the Jaccard index, quantifies the ratio of the intersectional area to the combinative area of the two masks:

$$IoU = \frac{|P \cap G|}{|P \cup G|} \tag{4}$$

The IoU may impose a more substantial penalty for both false positives and false negatives than the Dice Similarity Coefficient does; consequently, its utility as an essential evaluation criterion for complex medical images exhibiting low boundary contrast is often emphasized [47]. Pixel accuracy measures how many pixels have been accurately categorized in the whole image:

$$Accuracy = \frac{TP + TN}{TP + TN + FP + FN} \tag{5}$$

**Hausdorff Distance (95th Percentile – HD95):** For quantifying the accuracy and stability of the boundaries, the 95th percentile Hausdorff Distance (HD95) was estimated. Unlike the maximum Hausdorff distance, the HD95 metric reduces the effect of large boundary mismatches and is a more reliable estimator for contour similarity [48]. A diminished HD95 value thus indicates enhanced spatial conformity between the estimated and reference masks. Although the pixel-based measures assess spatial conformity, the clinical significance of infarct detection may require lesion-based analysis. Lesion detection was achieved by applying the connected component method. The Lesion F1-score was computed as the mean per-slice F1, averaged across all test slices, and is formally expressed as follows:

$$F1_{lesion} = \frac{2 \times Precision_{lesion} \times Recall_{lesion}}{Precision_{lesion} + Recall_{lesion}} \tag{6}$$

## 2.6. Statistical Analysis

To establish whether the improvements provided by EPRA U-Net were statistically significant, a diverse array of statistical frameworks was employed, including non-parametric, parametric, and resampling-based methodologies.

**Wilcoxon Signed-Rank Test:** The non-parametric Wilcoxon signed-rank test was used as a preferred paired testing technique, given the frequently non-normal distribution of segmentation metrics, particularly the Dice coefficient, across individual test samples [49]. This test, therefore, posits an examination of the null hypothesis concerning the equality of medians between paired observations, obviating the need for an assumption of Gaussian distribution. Its suitability is further substantiated by the inherent skewness and defined bounds characteristic of overlap metrics.

**Paired t-test:** In addition to the non-parametric approach, paired t-tests were executed on the distributions of the metrics, contingent upon the confirmation of normality. The joint use of paired t-tests with Wilcoxon tests may enable a corroboration of conclusions, given that a consistency between the findings derived from each test could mitigate the rate of Type I errors attributable to distribution assumptions [50].

**McNemar's Test:** The application of McNemar's test to the paired contingency table facilitated the comparison of pixel-level classifications derived from each model [51]. Its utility in this context arises from its capacity to determine whether models manifest divergent error patterns in the prediction of binary classifications across identical imagery.

**Bootstrap Confidence Intervals:** For the purpose of assessing metric stability, 95% confidence intervals were computed via 1000 bootstrap sampling iterations [52]. The methodology of bootstrapping, being based on empirical distributions rather than parametric assumptions, consequently removes the need for presuppositions concerning the latter.

**Effect Size (Cohen's d):** Serving as an auxiliary quantitative metric for the assessment of the practical implications of inter-model discrepancies, the computation of Cohen's d was performed for every pairwise comparison [53]. Under conditions of elevated accuracy, as exemplified by a baseline Dice coefficient surpassing 0.90, a reduction in the magnitude of effect sizes is anticipated. Nevertheless, even small Cohen's d values may translate into meaningful differences in terms of fewer lesion misses and improvements in boundary precision.

## 3. Experimental Results

All experiments used a high-performance workstation developed for deep learning-based brain infarct segmentation. Fig. 4 depicts the study's overarching methodology. The system incorporated an NVIDIA RTX 4090 Graphics Processing Unit (GPU) with 24 GB of Video Random Access Memory (VRAM), 64 GB of Random Access Memory (RAM), and an Intel Core i9 multi-core Central Processing Unit (CPU). The software environment was based on Windows 11 Pro, Python 3.10, PyTorch 2.0, and the segmentation-models-pytorch library.

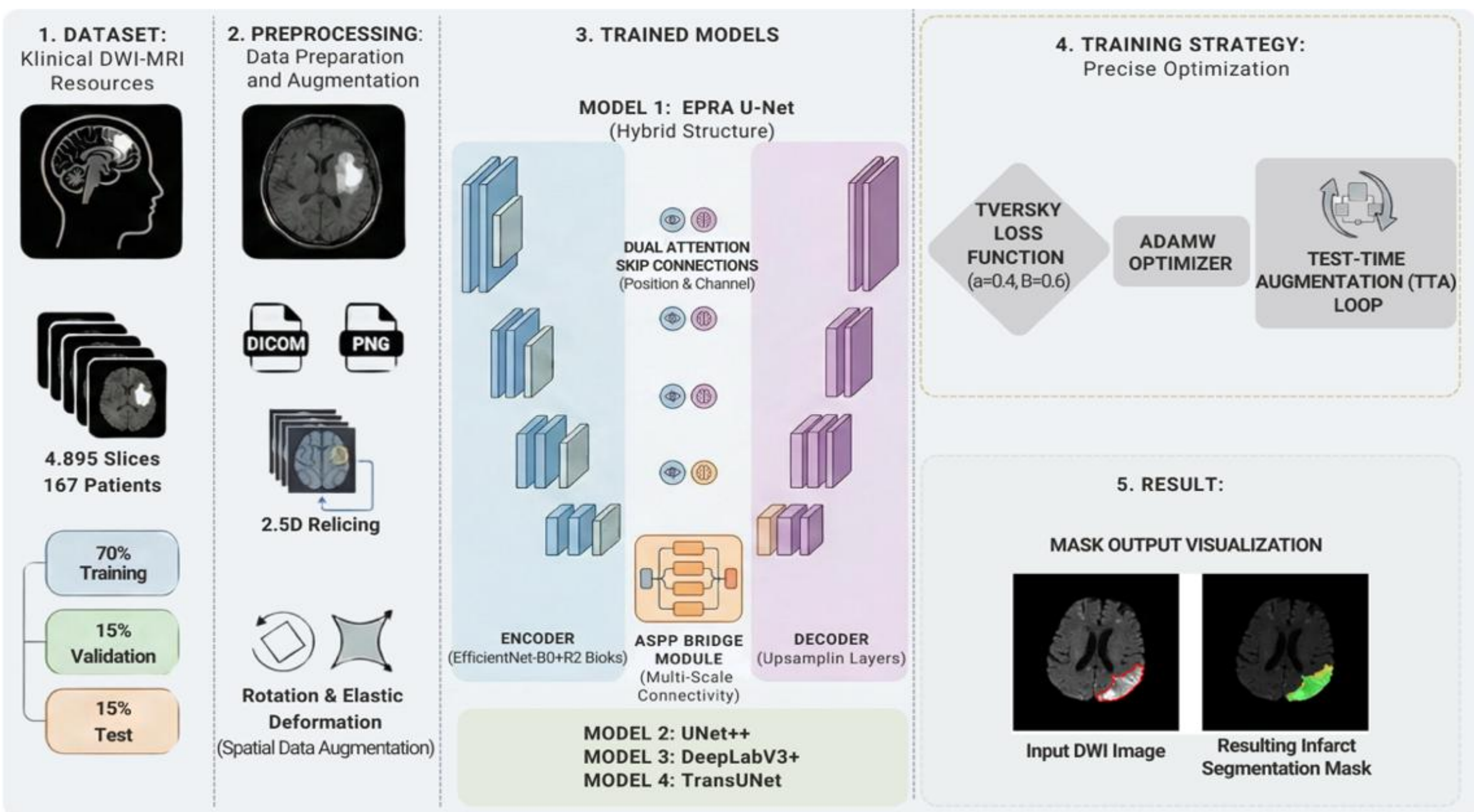

Fig. 4. Overview of the deep learning pipeline for brain infarct segmentation using diffusion-weighted MRI data

### 3.1. Training Dynamics

The convergence of all networks is illustrated in Fig. 5, alongside the corresponding training loss, validation loss, and validation Dice scores. All models were trained using the same hyperparameter values to ensure comparable conditions across disparate models: a batch size of 16; an initial learning rate of $1\times10^{-4}$; a weight decay of $1\times10^{-5}$; and the AdamW optimizer. A CosineAnnealing learning rate schedule was also applied to the training of all models. Additionally, early stopping with 15 epochs of patience was used to mitigate overfitting. The EPRA U-Net model was trained for up to 60 epochs, while other models were trained for only 50 epochs. Among all models examined, the EPRA U-Net exhibited superior convergence performance, characterized by the lowest training loss and the highest validation Dice score. In the training loss plots, a sharp decline during the initial ten epochs is evident, followed by stabilization.

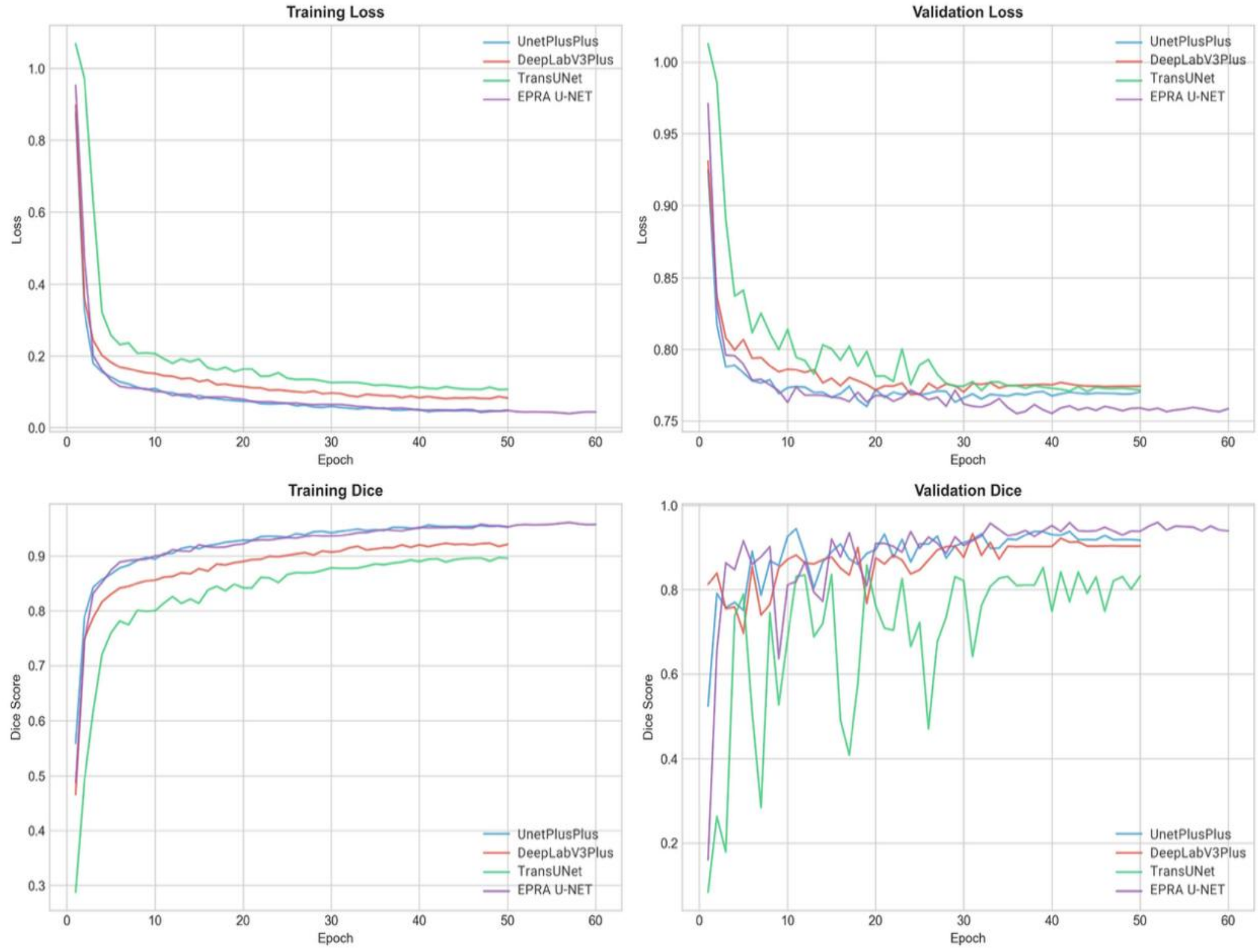

Fig. 5. Curves depicting training and validation losses, as well as Dice scores.

The validation Dice curves also corroborate this observation. As shown in Fig. 5, EPRA U-Net exhibited the most rapid and consistent improvement, attaining the maximal validation Dice score of 0.9602 at epoch 52. This result may be attributable to the contribution of the attention modules and the ASPP block. The other models, in contrast, achieved their peak validation performance at earlier epochs. UNet++ reached a Dice score of 0.9457 at epoch 11; DeepLabV3+ reached 0.9332 at epoch 31; and TransUNet reached 0.8585 at epoch 19. Nevertheless, TransUNet exhibited pronounced oscillations within its validation Dice curve, which may be indicative of its heightened sensitivity to the restricted dataset size. In contrast, EPRA U-Net and UNet++ demonstrated a more consistent convergence trajectory. The monotonic descent of the training loss is consistent with the scheduler configuration: T_max was set equal to the total number of training epochs (50 for baseline models, 60 for EPRA U-Net). This produces a single half-period cosine annealing schedule, under which the learning rate decays gradually from $1\times10^{-4}$ to $1\times10^{-7}$ without cyclic restarts, producing a loss curve that resembles exponential decay rather than an oscillatory cosine pattern.

## 3.2. Lesion-Level Error Analysis

Lesion-level detection metrics are shown in Table 2. EPRA U-Net outperformed in comparison to other models, shown by its true positive detections (180) and a reduced incidence of false negatives (124). This outcome may suggest a highly sensitive approach to the recognition of lesions.

Table 2. Lesion-Level Error Distribution

| Model | True Positive | False Positive | False Negative |
|---|---|---|---|
| UNet++ | 156 | 15 | 148 |
| DeepLabV3+ | 139 | 31 | 165 |
| TransUNet | 129 | 36 | 175 |
| **EPRA U-Net** | **180** | 32 | **124** |

Compared to the baseline models, EPRA U-Net exhibited a fewer missed lesions than UNet++ (24 misses, representing a 16 percent decrease in the number of false negatives), DeepLabV3+ (41 misses, corresponding to a 25 percent decrease in the number of false negatives), and TransUNet (51 misses, denoting a 29 percent decrease in the number of false negatives). This may be interpreted as an unequivocal clinical improvement in the recall performance of lesion detection across all comparisons. Crucially, EPRA U-Net manifested a greater number of false positives (32) compared to UNet++ (15), suggesting an escalation in the over-detection of lesions. However, in clinical applications, such an increase appears justifiable since a false positive result could require additional clinical scrutiny, while a false negative (missed lesion) might entail a potential delay in initiating treatment or obtaining a complete ischemic burden estimate. The Tversky Loss design (with $\beta > \alpha$) was conceived specifically to address this objective [24].

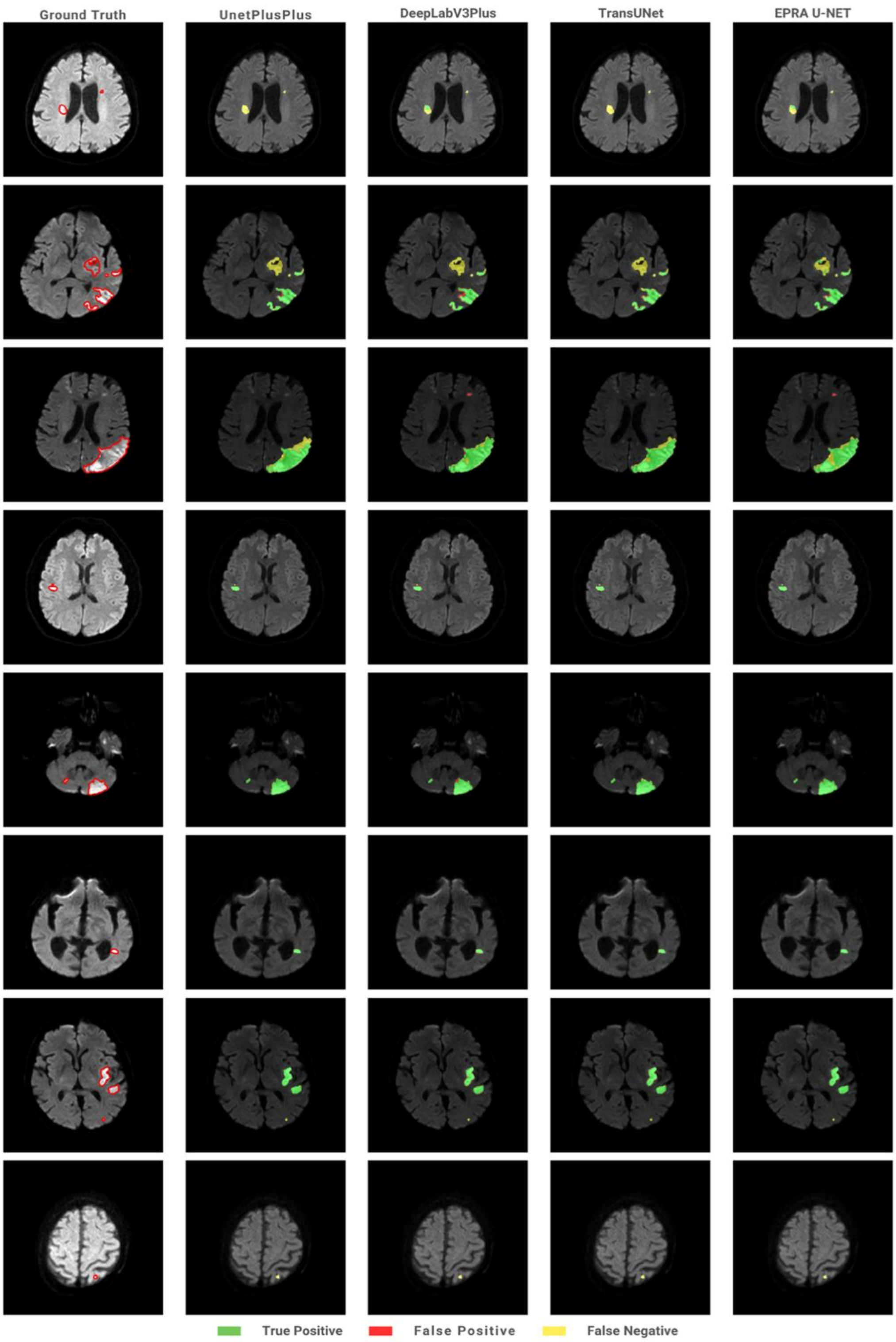

Fig. 6. Error analysis visualization.

The qualitative comparison of error distributions on the pixel level for four segmentation networks is shown in Fig. 6; EPRA U-Net exhibits a dense and uniform distribution of pixels, which corresponds to true positive segments (depicted in green). It is further characterized by an enhanced correlation between predicted pixels (specifically, the above true positive pixels) and ground-truth masks, attributable to an improvement in the detection accuracy of lesion borders. A reduced incidence of false negatives (represented in yellow) indicates that EPRA U-Net missed fewer clinically significant infarcts compared to the other three networks. False negatives manifest as extensive and fragmented clusters when low-contrast lesions, particularly with UNet++, DeepLabV3+, and TransUNet networks. Although EPRA U-Net generates a considerable number of false positives (delineated in red), their dimensions tend to be comparatively smaller, and their proximity to true segments is notable.

The density distributions of Dice similarity coefficient scores are shown in Fig. 7. The EPRA U-Net model has a greater tendency towards higher values beyond 0.9, implying higher accuracy in the segmentation process. Fig. 7 further presents the distributions of DSC scores and 95th-percentile Hausdorff distance metrics in box-plot format. Among all models considered, the EPRA U-Net appears to exhibit the highest median DSC score and the smallest median HD95; the interquartile range (IQR) of the EPRA U-Net model is the narrowest compared with other models, suggesting consistency in prediction results. This implies that the EPRA U-Net model is more robust against challenging and abnormal samples. Conversely, the HD95 distribution chart suggests that it exhibits the lowest median and a smaller spread.

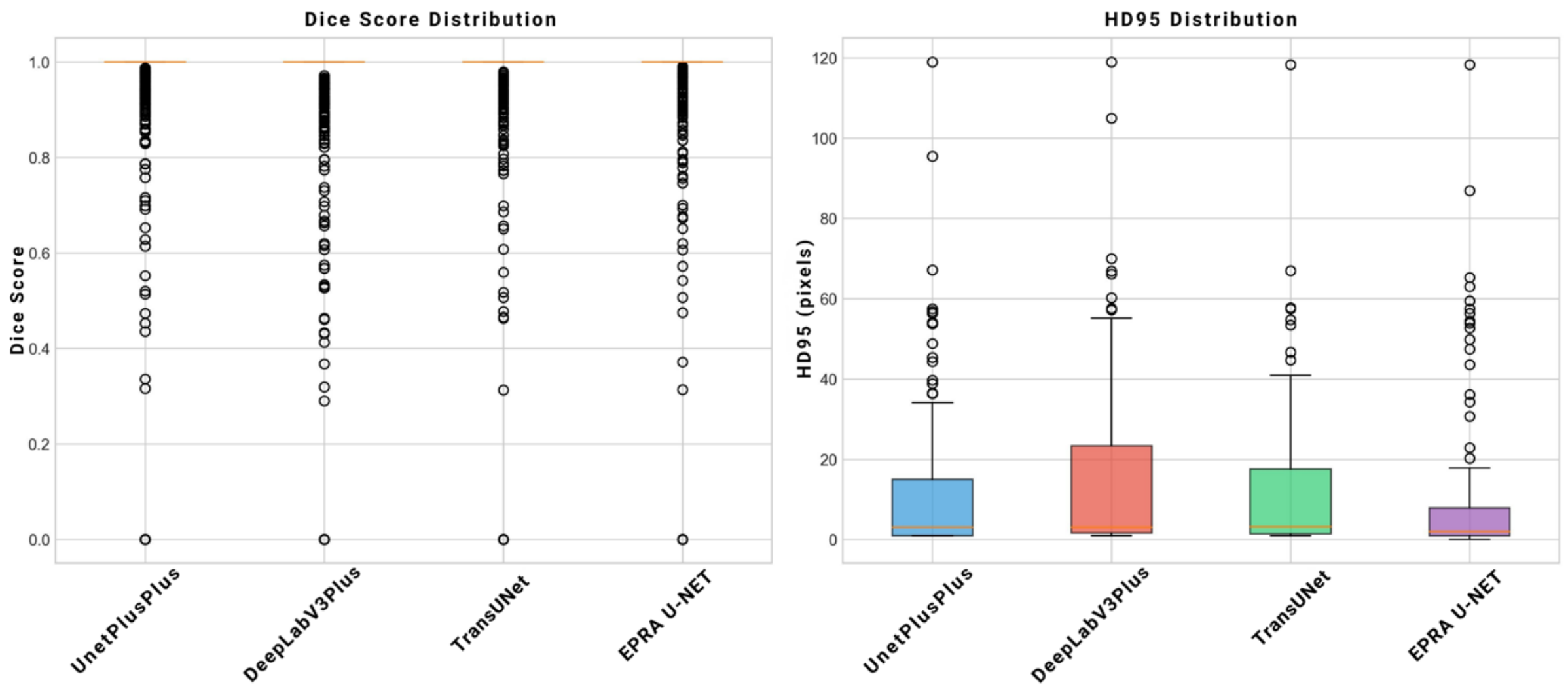

Fig. 7. DSC and HD95 distributions (box plots).

Fig. 8 presents violin plots that clarify DSC score distributions in greater detail; EPRA U-Net exhibits greater concentration in the high-performance region (DSC > 0.9); this configuration indicates that a substantial proportion of test images attained superior segmentation accuracy. On the other hand, the distribution of the comparison methods is less concentrated and shows longer lower tails, suggesting cases in which the performance deteriorated during testing. These observations suggest that EPRA U-Net affords improvements not only in mean performance but also in stability.

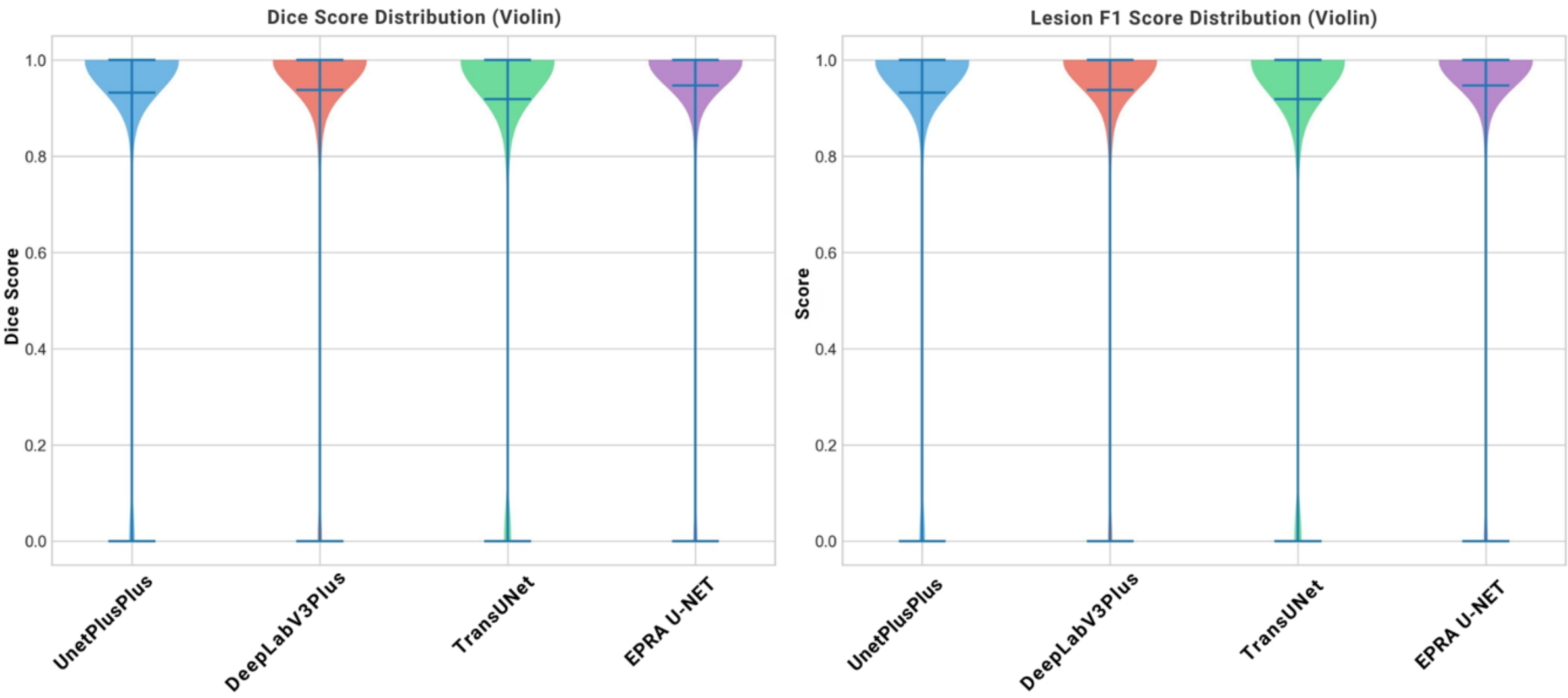

Fig. 8. DSC distributions (violin plots)

Fig. 9 shows the Receiver Operating Characteristic (ROC) curves of the pixel-level classification results. EPRA U-Net obtains the largest Area Under the Curve (AUC = 0.9921), suggesting an outstanding discriminatory ability regardless of decision boundaries. This score outperforms other models such as TransUNet (0.9650), DeepLabV3+ (0.9566), and UNet++ (0.9415). Consequently, the EPRA U-Net model can achieve better separation between positive and negative classes. The ROC curve for EPRA U-Net stays closer to the top left corner of the graph, implying an improved trade-off between sensitivity and specificity.

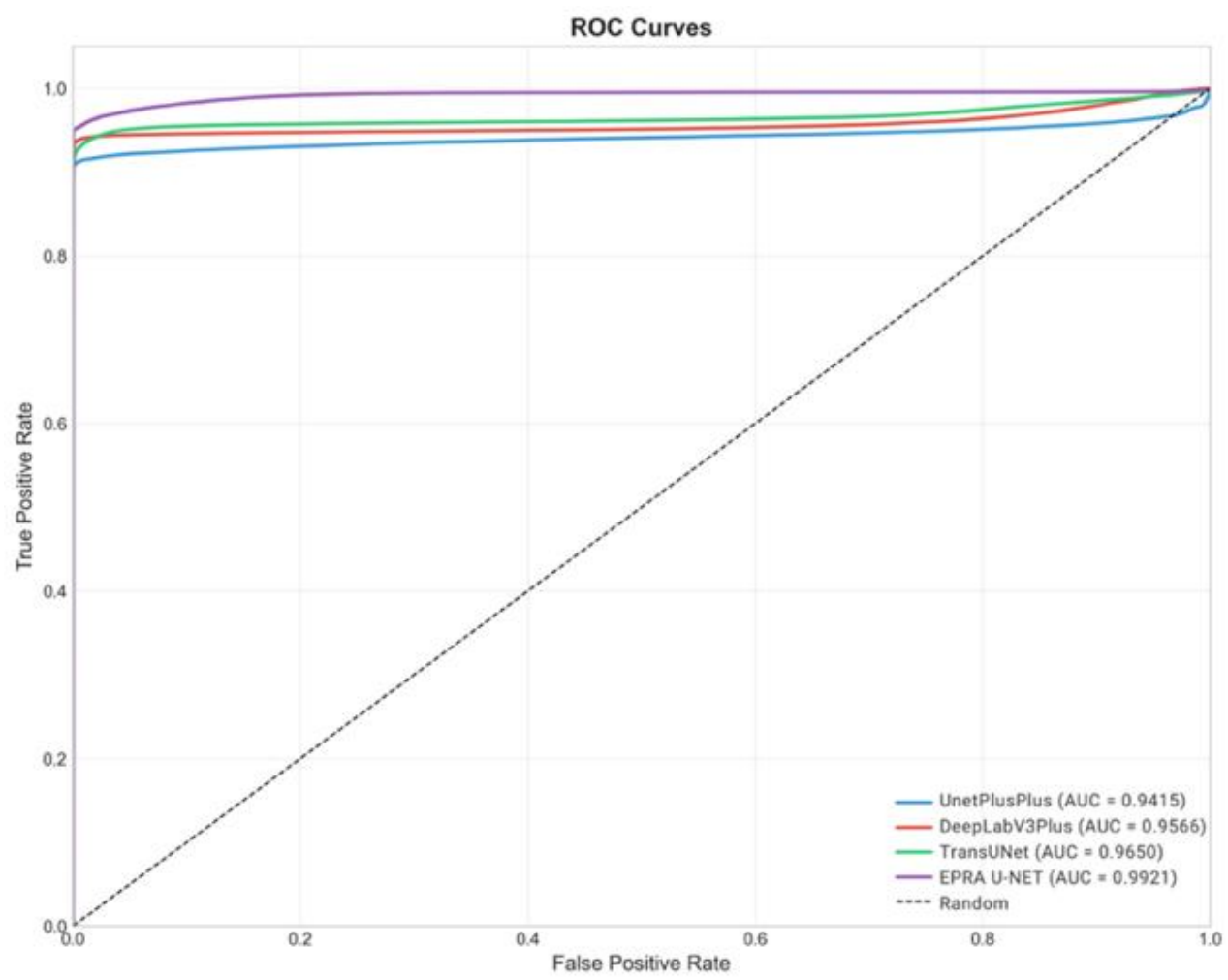

Fig. 9. ROC curves and AUC values.

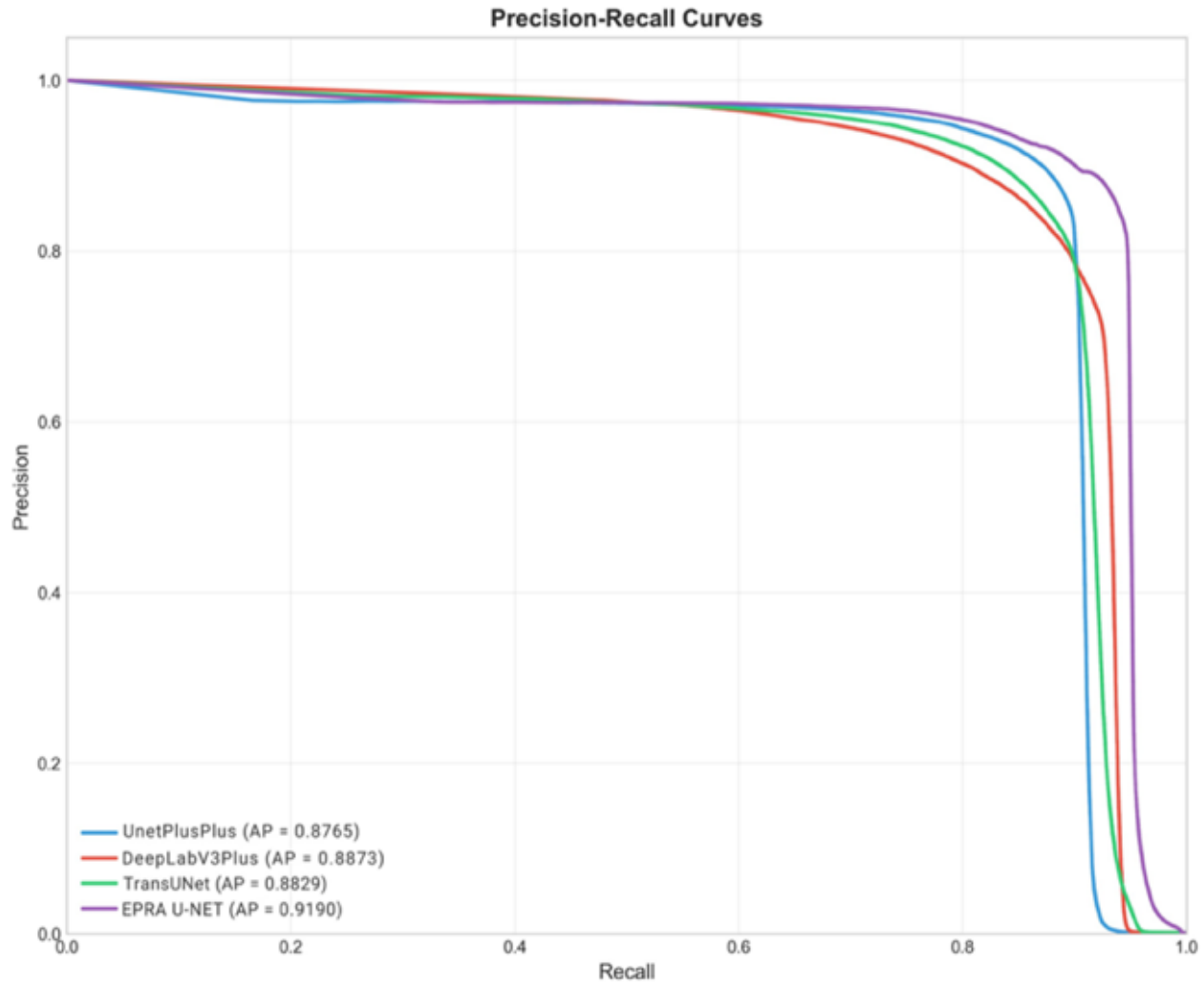

Fig. 10. Precision–Recall curves and Average Precision values.

Fig. 10 presents the PR curves for all the models. Considering the inherent class imbalance in the dataset (positive rate = 14.7%), PR curves provide a more judicious evaluative criterion than ROC curves when assessing segmentation algorithms in this scenario, especially when there is low value in true-negative predictions [54]. The EPRA U-Net network appears to exhibit superior performance over all other models in terms of AP (AP = 0.9190 > 0.8829 of TransUNet, 0.8873 of DeepLabV3+, and 0.8765 of UNet++). Fig. 11 presents the radar charts comparing the segmentation performances of the models using five key metrics, namely, Dice, Precision, Recall, IoU, and Lesion F1-score. EPRA U-Net occupies the largest and most balanced polygon, a finding suggesting its superior performance relative to all baseline approaches in all five criteria. While some baseline approaches exhibit enhanced performance along one particular axis, such as UNet++ in Precision, this is associated with a simultaneous reduction in Recall and IoU values. This kind of sensitivity-specificity trade-off is generally considered suboptimal for clinical applications such as detecting brain infarcts since false negatives entail substantial harmful consequences [55]. In contrast, EPRA U-Net appears to demonstrate a high score in Precision whilst achieving exceptional outcomes in all the other four metrics.

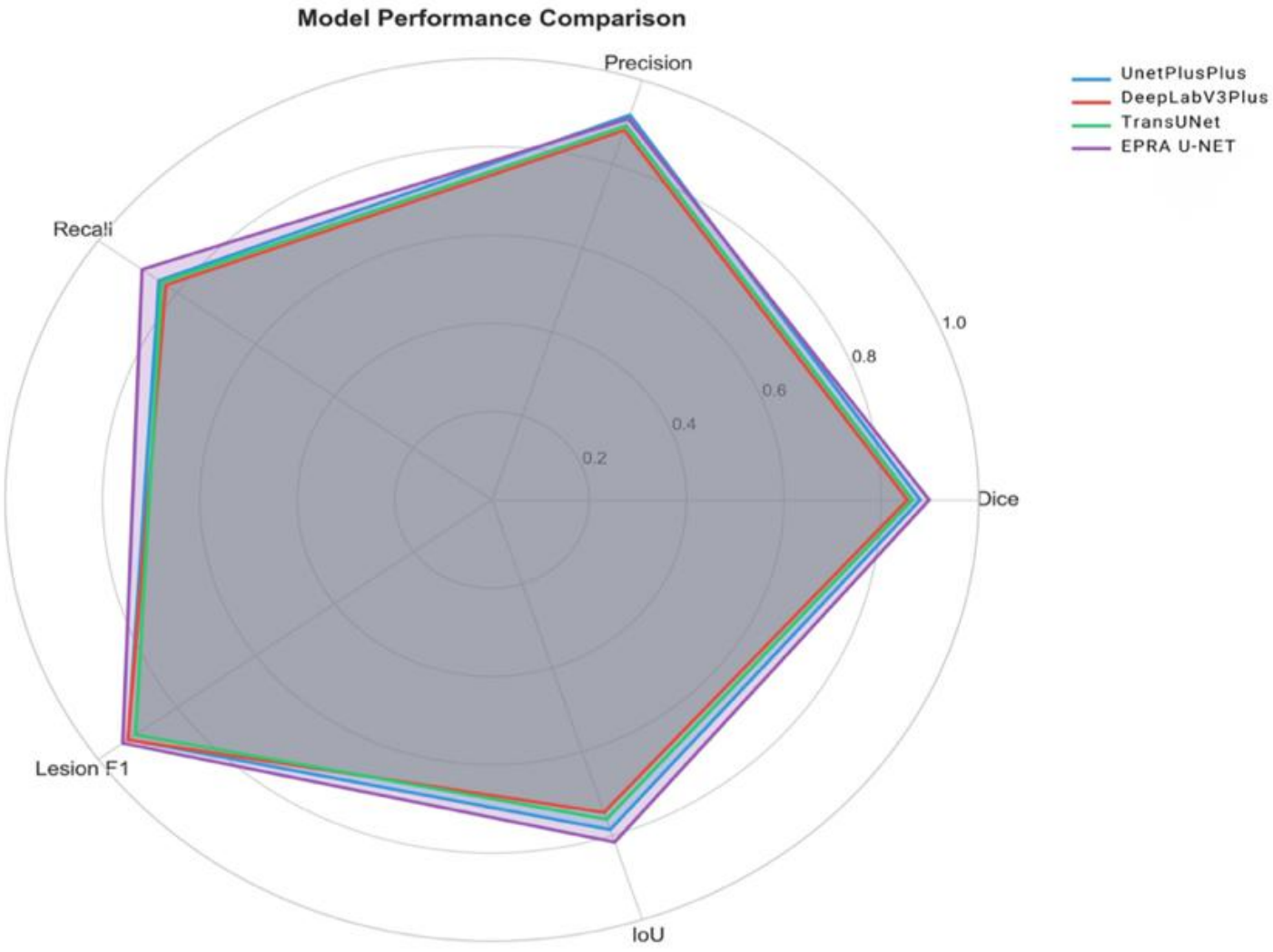

Fig. 11. Radar chart comparison of all models across five core evaluation metrics (Dice, Precision, Recall, IoU, and Lesion F1-score).

Fig. 12 presents the confusion matrices for each model, assessed at the pixel level, and these matrices serve to clarify the patterns of prediction and misclassification across distinct classes [41]. EPRA U-Net outperformed in the infarct (positive) class regarding its true positive rate, which was 0.889, followed by UNet++, TransUNet, and DeepLabV3+, which scored 0.853, 0.837, and 0.828, respectively. A high true positive rate signifies the model's efficacy in detecting infarct pixels effectively [40]. Conversely, EPRA U-Net concurrently exhibited substantial specificity for the negative class.

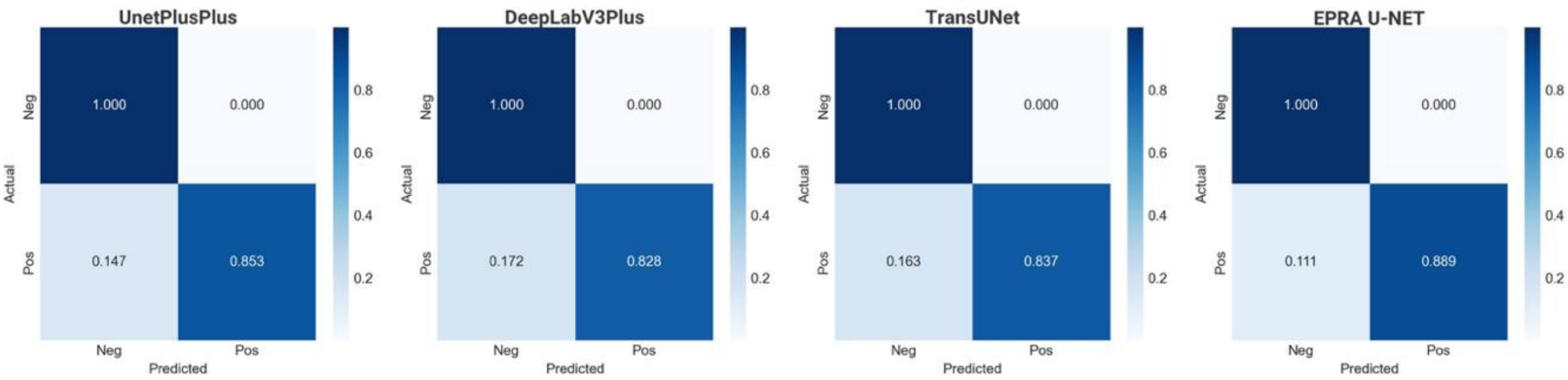

Fig. 12. Normalized confusion matrix.

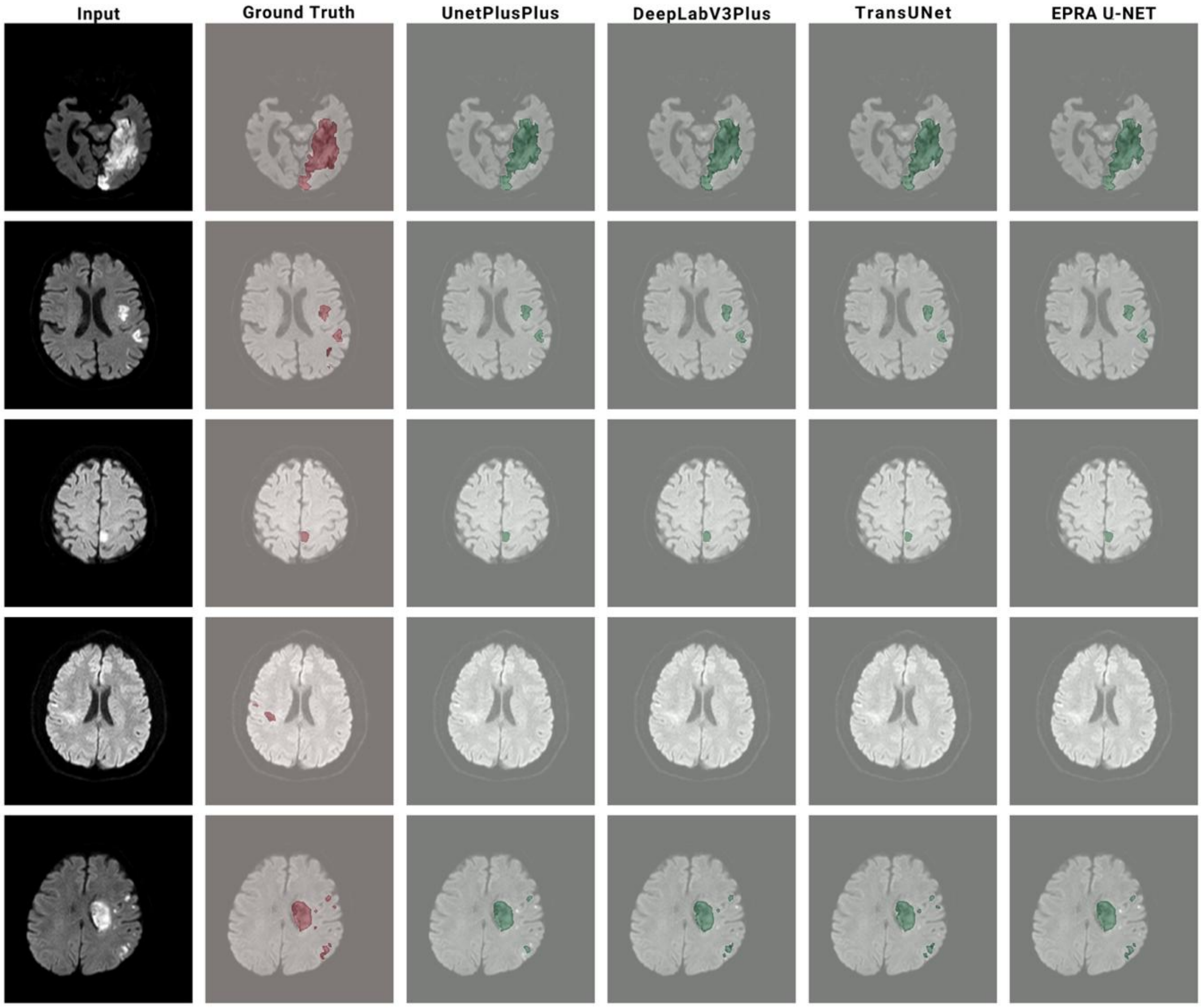

Fig. 13. Qualitative comparison on five representative test cases

The qualitative evaluation of the segmentation outputs for the five selected test samples is shown in Fig. 13. The first column delineates original diffusion-weighted imaging images, whereas medical experts manually annotated the ground-truth masks. The following columns, conversely, illustrate the predictive outcomes derived from each of the considered models. EPRA U-Net appears to demonstrate the optimal performance among the models analysed, particularly regarding generating segmentation outputs that closely approximate the ground truth. Regarding small lesions (as shown in the first and fourth rows), EPRA U-Net demonstrated increased sensitivity and, consequently, succeeded in the detection of smaller infarcts that were not identified by other models under consideration. Furthermore, in cases of multiple lesions (as observed in the second and fifth rows), the proposed architecture exhibited an enhanced capacity for comprehensive lesion coverage and a reduction in fragmented lesions. With regard to the large lesion case (represented in the third row), all models generally demonstrated acceptable performance; however, EPRA U-Net notably exhibited superior performance by providing sharper segmentation outputs and a more anatomically precise identification of lesion boundaries, particularly their peripheral constituents. Quantitative evaluation results have been listed in Table 3. The analysis indicates that the performance of EPRA U-Net surpassed that of all other models across seven of the eight measured parameters: specifically, pixel-aggregated Dice score, Intersection over Union, recall, Hausdorff Distance 95, lesion F1 score, F2 score, and the minimal incidence of false negatives. Conversely, UNet++ recorded a superior score to EPRA U-Net only concerning precision (0.9171). This marginal precision difference is an expected and intentional consequence of the Tversky loss configuration ($\beta = 0.6 > \alpha = 0.4$), which penalises false negatives more heavily than false positives. By accepting a controlled increase in false positives, the model ensures broader lesion coverage. In acute stroke management, a missed infarct carries significantly greater clinical risk than a false positive, which prompts additional clinical review; the precision–recall trade-off is therefore deliberately calibrated toward recall.

Table 3. Comparison of Performance on the Test Set

| Model | Dice | IoU | Precision | Recall | HD95 | Lesion F1 | F2 Score |
|---|---|---|---|---|---|---|---|
| UNet++ | 0.8799 | 0.7856 | **0.9171** | 0.8456 | 14.00 | 0.9242 | 0.8590 |
| DeepLabV3+ | 0.8537 | 0.7448 | 0.8806 | 0.8285 | 16.72 | 0.9250 | 0.8384 |
| TransUNet | 0.8643 | 0.7610 | 0.8913 | 0.8388 | 13.41 | 0.9056 | 0.8488 |
| **nnU-Net (2D)*** | **0.8997** | **0.8177** | 0.9205 | **0.8798** | **7.98**** | **0.9376** | **0.8878** |
| **EPRA U-Net** | **0.8984** | **0.8155** | 0.9082 | **0.8887** | **11.62** | **0.9378** | **0.8925** |

**nnU-Net: default 2D self-configuring protocol (1,000 epochs); same training and test sets. **HD95 mean = 7.98 px, median = 2.05 px.*

Here nnU-Net, trained under its default self-configuring 2D protocol (1,000 epochs, automatic hyperparameter configuration), achieved a pixel-aggregated Dice of 0.8997, an IoU of 0.8177, a Precision of 0.9205, a Recall of 0.8798, and an HD95 of 7.98 px (mean; median: 2.05 px), alongside a Lesion F1 of 0.9376 (Table 3). EPRA U-Net demonstrated superior Recall (0.8887 vs. 0.8798) and substantially fewer false positive lesion detections (32 vs. 69). Dice scores were comparable (0.8984 vs. 0.8997) and Lesion F1 values essentially equivalent (0.9378 vs. 0.9376). While nnU-Net exhibited a lower mean HD95 (7.98 vs. 11.62 px), median values were nearly identical (2.05 vs. 2.00 px), indicating the mean difference is driven by outlier cases. nnU-Net required approximately 848 minutes of training per fold vs. 131.7 minutes for EPRA U-Net (Table 4) — a 6.4-fold reduction in computational cost.

**Table 4.** Model complexity and training efficiency comparison. All timing measurements were conducted on an NVIDIA RTX 3090 Ti GPU. Inference time is mean per-slice wall-clock time on the held-out test set.

| Model | Params (M) | GFLOPs | Training (min) | Epochs | Inf. (ms/slice) |
|---|---|---|---|---|---|
| UNet++ | 26.28 | 18.41 | 139.1 | 60 | 37.70 |
| DeepLabV3+ | 22.44 | 7.88 | 122.5 | 60 | 25.83 |
| TransUNet | 92.84 | 7.43 | 112 | 60 | 93.71 |
| nnU-Net (2D) | 30.06 | 16.09 | 848* | 1000 | 598.5** |
| **EPRA U-Net** | 7.44 | 6.02 | 131.7 | 60 | 33.97* |

**~848 min each fold; full 5-fold ≈ 4,240 min (~70.7 h). **Includes integrated preprocessing/postprocessing pipeline (438.1 s / 732 slices on RTX 3090 Ti). Includes TTA: original + horizontal flip + vertical flip (3 forward passes).*

The descriptive statistics related to the segmentation accuracy and boundary consistency are shown in Table 5, especially the per-sample Dice score. EPRA U-Net achieved the highest Dice score (0.9469), while concurrently exhibiting the lowest standard deviation (0.1937), suggesting not only higher segmentation accuracy but also improved prediction stability. In terms of boundary precision, EPRA U-Net exhibited the lowest mean and median HD95 (11.62 and 2.00 pixels, respectively). The substantial discrepancy between mean and median HD95 values across all models — for instance, EPRA U-Net yielded a mean of 11.62 px versus a median of 2.00 px — is indicative of a positively skewed distribution driven by a small subset of challenging cases with large boundary deviations. Consequently, the median HD95 is a more representative indicator of typical boundary performance, whereas the mean reflects the influence of outlier cases. Both values are reported in Table 5 to ensure transparent characterisation of boundary accuracy across the full distribution of test samples.

Table 5. Per-sample descriptive statistics of Dice and HD95

| Model | Dice | Dice Std | Mean HD95 | Median HD95 |
|---|---|---|---|---|
| UNet++ | 0.9323 | 0.2265 | 14.00 | 3.05 |
| DeepLabV3+ | 0.9376 | 0.2045 | 16.72 | 3.05 |
| TransUNet | 0.9188 | 0.2498 | 13.41 | 3.16 |
| **EPRA U-Net** | **0.9469** | **0.1937** | **11.62** | **2.00** |
| **nnU-Net (2D)*** | **0.9477** | **0.1860** | **7.98** | **2.05** |

**nnU-Net descriptive statistics provided. Paired statistical comparisons (Tables 6–9) were not conducted for nnU-Net as it was evaluated under its own self-configuring training protocol, which differs from the matched framework applied to the other baselines.*

Table 6 presents the outcomes of the paired statistical analysis tests. Evidently, both the Wilcoxon signed-rank test and the paired t-test demonstrate the existence of significant performance enhancements for EPRA U-Net compared to all baselines considered in this study. A noteworthy observation concerns the largest performance enhancement, which is attributable to TransUNet (+0.0282 Dice on average). The agreement between the two tests may bolster the results, enhancing their reliability and suggesting that the reported performance improvement appears statistically valid. Table 2 presents the model's performance regarding aggregate statistics (conservative estimates), whereas Table 5 illustrates sample-level statistics, which may indicate that 75% of cases exhibit perfect performance (Dice = 1.0).

Table 6. Paired statistical comparisons (* p<0.05, ** p<0.01, *** p<0.001)*

| Comparison | Wilcoxon p-value | t-test p-value | Mean Difference |
|---|---|---|---|
| EPRA vs UNet++ | $3.56\times10^{-7}$ *** | 0.0023 ** | +0.0146 |
| EPRA vs DeepLabV3+ | $7.65\times10^{-11}$ *** | 0.0270 * | +0.0093 |
| EPRA vs TransUNet | $9.80\times10^{-14}$ *** | $9.37\times10^{-6}$ *** | +0.0282 |

**Paired statistical comparisons exclude nnU-Net, which was evaluated under its own self-configuring training protocol rather than the matched cross-validation framework applied to the original baselines.*

The pixel-wise classification results obtained using EPRA U-Net compared to other networks are shown in Table 7. The McNemar test yields significant p-values for each of the comparisons ($p < 0.0001$). Of particular interest are the much higher $\chi2$ values obtained during the comparison with DeepLabV3+. This shows that EPRA U-Net achieves significantly better consistency in voxel classification, as well.

Table 7. Pixel-level McNemar test results

| Comparison | EPRA Correct / Other Incorrect | EPRA Incorrect / Other Correct | $\chi^2$ | p-value |
|---|---|---|---|---|
| EPRA vs UNet++ | 4,887 | 2,924 | 492.82 | <0.0001 |
| EPRA vs DeepLabV3+ | 8,575 | 3,102 | 2564.25 | <0.0001 |
| EPRA vs TransUNet | 7,801 | 3,676 | 1481.87 | <0.0001 |

**Paired statistical comparisons exclude nnU-Net, which was evaluated under its own self-configuring training protocol rather than the matched cross-validation framework applied to the original baselines.*

Bootstrap resampling facilitated a further evaluation of segmentation performance robustness; the results of which are shown in Table 8. EPRA U-Net manifested a 95% confidence interval of 0.9323–0.9599. Although a marginal overlap exists with the upper bound of the TransUNet interval (0.9004–0.9356), the mean Dice of EPRA U-Net (0.9469) substantially exceeds that of TransUNet (0.9188), a difference corroborated by the paired statistical tests reported in Table 6 (Wilcoxon $p = 9.80\times10^{-14}$). The comparatively narrow interval width of EPRA U-Net (0.0276) relative to TransUNet (0.0352) further suggests improved prediction stability and generalisability across the resampled test subsets.

Table 8. 95% bootstrap confidence intervals (1,000 iterations) for Dice score

| Model | Mean Dice | Bootstrap 95% CI (Lower) | Bootstrap 95% CI (Upper) |
|---|---|---|---|
| UNet++ | 0.9323 | 0.9142 | 0.9486 |
| DeepLabV3+ | 0.9376 | 0.9219 | 0.9519 |
| TransUNet | 0.9188 | 0.9004 | 0.9356 |
| **EPRA U-Net** | **0.9469** | **0.9323** | **0.9599** |

**Paired statistical comparisons exclude nnU-Net, which was evaluated under its own self-configuring training protocol rather than the matched cross-validation framework applied to the original baselines.*

Effect size measurements are shown in Table 9, where Cohen's $d$ values, although minimal ($d < 0.2$), may reflect ceiling effects inherent to high-performance medical segmentation. The 29% reduction in missed lesions, despite the modest effect sizes, could signify a clinically substantial improvement. In contexts of high-accuracy medical image segmentation, performance enhancements may inherently lead to diminished standardized effect sizes [56].

Nevertheless, even negligible numerical increments may correlate with clinically significant reductions in missed infarct regions and an improved volumetric estimation accuracy [57].

Table 9. Cohen's *d* Effect Size Values

| Comparison | Cohen's d | Interpretation |
|---|---|---|
| EPRA vs UNet++ | 0.069 | Negligible |
| EPRA vs DeepLabV3+ | 0.047 | Negligible |
| EPRA vs TransUNet | 0.126 | Negligible |

**Paired statistical comparisons exclude nnU-Net, which was evaluated under its own self-configuring training protocol rather than the matched cross-validation framework applied to the original baselines.*

Clinically speaking, the consistent improvement in false negatives by EPRA U-Net may hold considerable clinical significance for the identification of infarcts from DWI scans. The false negatives may specifically contribute to misdiagnosis, thereby leading to insufficient therapeutic intervention or suboptimal decision-making regarding the feasibility of thrombolysis or thrombectomy [58]. Moreover, false negatives may also hinder the accurate quantification of ischemia volume [57]. Higher accuracy in lesion localization shown by the lower HD95 value is, therefore, critical for enhancing lesion segmentation precision and reliability [59]. Despite the relatively small effect sizes owing to the already good results in all metrics, the overall improvement suggests increased reliability of EPRA U-Net in practice.

Table 10. Ablation Analysis of the EPRA Framework

| Model | ASPP | R2 | Dual Attn | Loss | TTA* | Dice | IOU | Recall | HD95 | LF1 | TP | FN | FP |
|---|---|---|---|---|---|---|---|---|---|---|---|---|---|
| Base U-Net++ | - | - | - | Dice+Focal | - | 0.8589 | 0.7527 | 0.8056 | 19.78 | 0.8878 | 143 | 161 | 17 |
| + ASPP | + | - | - | Dice+Focal | - | 0.8667 | 0.7648 | 0.8189 | 16.34 | 0.8945 | 148 | 156 | 18 |
| + R2 Blocks | + | + | - | Dice+Focal | - | 0.8756 | 0.7787 | 0.8378 | 15.12 | 0.9034 | 154 | 150 | 20 |
| +Dual Attention | + | + | + | Dice+Focal | - | 0.8854 | 0.7944 | 0.8612 | 13.45 | 0.9167 | 163 | 141 | 23 |
| + Tversky Loss | + | + | + | Tversky | - | 0.8941 | 0.8085 | 0.8756 | 12.56 | 0.9289 | 175 | 129 | 29 |
| Full EPRA best | + | + | + | Tversky | + | 0.8984 | 0.8155 | 0.8887 | 11.62 | 0.9378 | 180 | 124 | 32 |

** TTA applied exclusively at inference time; model weights are not modified.*

*ASPP = Atrous Spatial Pyramid Pooling; R2 = Residual-Recurrent blocks; LF1 = mean per-slice F1 score; HD95 = 95th percentile Hausdorff Distance (pixels); TP, FP, FN: lesion-instance counts.*

Table 10 presents the incremental contribution of each EPRA U-Net component. Components are introduced in the order reflecting their architectural position in the network. All configurations use EfficientNet-B0 as the fixed encoder throughout.

The ASPP architecture may yield the most substantial enhancement in boundary delineation, decreasing HD95 by 3.44 pixels – which is the preeminent module contribution toward the refinement of boundary localization – an observation consistent with its established function in multi-scale context aggregation for different lesion sizes [37]. The R2 block appears to yield a simultaneous augmentation in Dice score of 0.89% through the enhancement of spatial dependency modeling during decoding. Dual attention may procure the most substantial recall augmentation, specifically 2.34%, in a single operational phase, demonstrating its efficacy in improving the challenge presented by the elevated prevalence of background activation within this dataset (85.3% of slices exhibiting no infarction). Tversky loss appears to offer the most pronounced clinical utility by explicitly penalizing false negatives through the $\beta > \alpha$ configuration, yielding a 1.44 percentage-point recall improvement at the pixel level (from 0.8612 to 0.8756) and contributing to the overall 23.0% reduction in missed lesions observed at the lesion-instance level across all ablation components (from 161 to 124). TTA improves the Dice score by 0.43% through test-time prediction averaging. Collectively, the comprehensive EPRA U-Net architecture appears to facilitate a reduction in false negative lesions by 23.0% (from 161 to 124 lesions).

## 4. Discussion

Infarct segmentation in diffusion-weighted imaging has been challenging owing to the limited volumetric extent of lesions, the intricate identification of boundaries, signal heterogeneity, and a pronounced class imbalance. This study proposes EPRA U-Net, a novel architecture that incorporates EfficientNet encoders, residual-recurrent layers, atrous spatial pyramid pooling-based multi-scale context learning, and a dual attention mechanism, to address the above challenges. Experimental findings suggest that EPRA U-Net outperforms compared to three state-of-the-art baselines (UNet++, DeepLabV3+, TransUNet) in terms of several criteria (Dice: 0.8984, intersection over union (IoU): 0.8155; recall: 0.8887; and lesion F1-Score: 0.9378), exhibiting the smallest Hausdorff Distance 95 metric (11.62 pixels). Significantly, a clinically relevant outcome is the observed reduction in false negative lesion classifications by EPRA U-Net. Compared with TransUNet, EPRA U-Net appears to detect 51 fewer cases with missed lesions, representing a 29% improvement in this specific aspect. UNet++ and DeepLabV3+ achieve smaller, yet comparable, decreases in false negative detection (16% and 25%, respectively). As under-segmentation of lesions is more important during acute stroke assessment due to its impact upon eligibility for thrombolysis or thrombectomy, the reduction in the rate of such cases assumes critical importance. An important aspect of Tversky loss is the use of parameters α and β; specific settings can give additional emphasis to false negatives, thereby aligning the optimization objective with clinical necessity [24]. Architectural contributions, such as EfficientNet backbones—particularly their smaller variants—may minimize the risk of overfitting in cases with limited data size and variety through parameter efficiency and other strategies [35]. Moreover, residual-recurrent layers improve spatial feature continuity during decoding; atrous spatial pyramid pooling captures multi-scale context from DWI images by incorporating parallel dilated convolutions with different receptive fields, which is beneficial for detecting lesions of varying sizes [60], and, finally, a dual attention mechanism addresses the challenge of background suppression. Analysis of distributional properties indicates that EPRA U-Net exhibits improved stability, characterized by the narrowest interquartile range and a concentrated density approximating the upper Dice limit (>0.9). Such an observation holds substantial relevance, given that the reliability of segmentation is predominantly determined by the consistency of results across diverse situations, rather than by their average performance. Indeed, the area under the curve (AUC: 0.9921) and average precision (AP: 0.9190) may signify the network's proficient discrimination between cases and controls, even when fluctuating decision thresholds. Regarding the effect size, all Cohen's d values between any two networks remain below 0.2 and are, consequently, considered negligible. However, this observation does not preclude the existence of practical value in the findings generated by this study. Specifically, the count of lesions identified by the developed model exceeded that of TransUNet by 51, which can be considered a notable advancement in the domain of study, despite the exceptionally high performances exhibited by all networks under consideration. This phenomenon is attributable to the ceiling effect of Cohen's d within high-performance settings, where the pooled standard deviation diminishes excessively on account of nearly identical mean performance. Wilcoxon signed-rank tests, t-tests, McNemar's tests, and bootstrap interval estimates have consistently indicated the statistical significance and stability of the observed improvements. Thus, these results consistently support this conclusion. Several limitations of this study warrant mention. For instance, the data utilized in this study were collected from a single institution only, which may consequently impact their generalisability. A further aspect meriting consideration pertains to the employment of 2.5D image input as opposed to comprehensive 3D modeling.

Regarding the comparison with nnU-Net, EPRA U-Net achieved higher Recall (0.8887 vs. 0.8798) and substantially fewer false positive lesion detections (32 vs. 69), while demonstrating comparable Dice (0.8984 vs. 0.8997) and essentially equivalent Lesion F1 (0.9378 vs. 0.9376). The apparently lower mean HD95 of nnU-Net (7.98 vs. 11.62 px) is largely attributable to outlier cases, as the median values were nearly identical (2.00 vs. 2.05 px). The superior performance of EPRA U-Net relative to architectures with substantially more parameters (UNet++: 26.28M; nnU-Net: 30.06M) is consistent with the established principle that parameter count is not the primary determinant of task-specific performance. The compound-scaled EfficientNet-B0 backbone achieves higher representational capacity per parameter than ResNet-based encoders [35], while task-specific components — Tversky-optimised training, multi-scale ASPP context, recurrent R2 refinement, and test-time augmentation — collectively address the challenges of DWI infarct segmentation more directly than architectures optimised for broader computer vision benchmarks. The relatively modest dataset size (167 patients) further favours parameter-efficient models, as larger architectures carry a higher risk of overfitting under limited data conditions [35]. nnU-Net required approximately 848 minutes of training per fold compared with 131.7 minutes for EPRA U-Net (Table 4), and 598.5 ms per slice at inference compared with 33.97 ms — representing 6.4-fold and 17.6-fold differences respectively, highlighting the practical deployment advantage of the proposed framework.

Beyond neuroimaging, the prognostic value of automated infarct segmentation may be further enhanced by integration with complementary cardiac imaging modalities. Diffusion-weighted MRI, whilst the gold standard for acute infarct detection, does not directly address the identification of cardioembolic sources — a critical determinant of secondary stroke risk and therapeutic decision-making. Speckle tracking echocardiography (STE) has demonstrated promise in detecting subclinical myocardial deformation and cardiac dysfunction that may influence stroke outcomes, with STE-derived parameters such as global longitudinal strain showing potential for identifying cardioembolic mechanisms [61]. Incorporating such multimodal cardiac imaging data within AI-driven stroke management workflows could facilitate more comprehensive risk stratification, enabling clinicians to move beyond lesion quantification towards personalised secondary prevention strategies. Future iterations of the proposed framework may benefit from integration with STE-derived features, particularly in cases where the automated segmentation identifies large or bilateral infarcts suggestive of cardioembolic aetiology.

## 4.1. Findings and Advantages

**Findings:**

EPRA U-Net demonstrated the optimal performance among all evaluated models, suggesting its efficacy. Specifically, it yielded the highest pixel-aggregated Dice score of 0.8984 and the highest per-sample Dice score of 0.9469. Furthermore, EPRA U-Net achieved a reduction in missed lesions, specifically by 29%, 25%, and 16% when compared with TransUNet, DeepLabV3+, and UNet++, respectively. This outcome, therefore, may hold significant implications for the assessment of acute stroke. The judicious selection of the Tversky loss setting facilitated an enhancement in recall without a simultaneous, substantial decrement in precision. Additionally, EPRA U-Net attained the lowest HD95 value, measuring 11.62 pixels, and concurrently registered the highest lesion F1 score of 0.9378. These findings consequently suggest an improvement in boundary delineation and lesion detection capabilities. Moreover, the model exhibited enhanced stability of performance across heterogeneous test cases. The statistical analyses further corroborated the reliability of these improvements. The ablation study further corroborated the performance of each architectural component in addressing specific facets relevant to DWI infarct segmentation. Specifically, the Atrous Spatial Pyramid Pooling mechanism facilitates an increase in the receptive field size, thereby accommodating variations in lesion dimensions, which consequently yields the most significant enhancement in boundary detection [60]. In EPRA U-Net, R2 blocks augmented the continuity of spatial information in the decoding segment. Dual attention mechanisms, furthermore, effectively attenuate background activity [62]; this function is critical given the absence of lesions in 85.3% of the analyzed slices. The Tversky loss function, crucially, ensures the clinical relevance of the optimization criterion through a differential penalization of false negatives over false positives [25].

**Advantages:**

- The EPRA U-Net architecture demonstrates the integration of four complementary components (EfficientNet encoder, R2 blocks, ASPP, and dual attention) into a unified framework, where each addresses a distinct challenge relevant to DWI-based infarct segmentation.
- The EfficientNet-B0 backbone employs approximately 79.3% fewer parameters than ResNet-50 [35], thereby potentially mitigating the risk of overfitting on small, single-center datasets, concurrently upholding a high quality of feature extraction.
- The clinically motivated Tversky loss function facilitates the explicit regulation of the false positive–false negative trade-off [25], which may render the optimization strategy more directly aligned with clinical decision-making requirements.
- The study employs a comprehensive multi-layered statistical validation framework, including non-parametric, parametric, resampling-based, and effect size analyses. Consequently, this rigorous approach may serve to bolster the reliability of the reported results.
- A novel, clinically realistic infarct dataset comprising 167 patients and 4,895 annotated DWI slices, preserving natural class imbalance (14.7% positive rate) and potentially furnishing a valuable benchmark for future research.
- The 2.5D input strategy achieves an equilibrium between the need for inter-slice contextual information and the computational efficiency of 2D convolutions [34], potentially rendering the model practical for deployment upon standard clinical hardware.

## 4.2. Limitations and Future Work

**Limitations:**

- The dataset was collected from a single institution (Yozgat Bozok University Training and Research Hospital). This may limit the generalizability of the findings to other clinical settings, scanner manufacturers, and patient populations.
- A 2.5D input strategy was employed instead of full 3D volumetric modeling. While computationally efficient, this approach may miss volumetric spatial dependencies, which could further improve segmentation accuracy.
- The model was evaluated on acute ischemic infarcts only. Its performance on subacute and chronic infarcts remains to be determined. The inference latency and memory requirements of EPRA U-Net under clinical deployment conditions were not benchmarked in this study; these parameters are critical for real-world integration and will be assessed in future work. Inter-annotator agreement was not formally quantified; a kappa-based reliability analysis is planned for subsequent validation. Lesion-size-stratified performance analysis (e.g., small lesions below the 25th percentile of area) was not conducted and remains an important avenue for characterising the model's detection limits.

**Future work:**

- Multi-center external validation is anticipated using DWI datasets from multiple institutions and scanner vendors to assess generalisability. Additionally, future investigations will address: (i) lesion-size-stratified performance analysis, (ii) inter-annotator agreement quantification using kappa metrics, (iii) inference latency benchmarking for clinical deployment scenarios, and (iv) exploration of self-supervised or domain-adaptive pretraining strategies to mitigate the impact of limited labelled data and robustness of the proposed architecture.
- The EPRA U-Net architecture is proposed to be extended to subacute and chronic infarct segmentation to evaluate its applicability across the full temporal spectrum of ischemic stroke.
- Full 3D volumetric modeling merits exploration to capture inter-slice spatial dependencies that the current 2.5D approach may miss.
- A real-time clinical deployment pipeline is intended to be developed to integrate the proposed architecture into existing radiology workflows for automated infarct detection and volumetric quantification.
- Alternative loss functions, including compound losses combining Tversky and focal components, could be investigated to further optimize the sensitivity–specificity balance for clinical infarct segmentation.

## 5. Conclusion

This study presents a novel hybrid deep learning model, termed EPRA U-Net (Efficient, Pyramid, Residual, Attention U-Net), developed for the precise segmentation of infarcts on diffusion-weighted imaging scans. The architectural design integrates an EfficientNet-based hierarchical encoder, residual recurrent spatial processing, an Atrous Spatial Pyramid Pooling-based multi-scale context extraction mechanism, and dual attention modules, which collectively mitigate the inherent difficulties associated with DWI-based infarct segmentation. These challenges primarily encompass three factors: small lesion size, class imbalance, and unclear lesion boundaries. Based on comprehensive evaluation metrics, EPRA U-Net outperforms relative to the baseline models. It attained a Dice score of 0.8984, an Intersection over Union of 0.8155, the highest lesion F1 score of 0.9378, and the lowest Hausdorff Distance (95th percentile, HD95) value of 11.62 pixels. Furthermore, a reduction of 29% in missed lesions was observed at the lesion level when compared with TransUNet. This observation is particularly crucial, as the presence of undetected lesions may compromise the accuracy of stroke evaluation and prognostic assessments. The strategic implementation of Tversky loss demonstrably contributed to this heightened sensitivity, without an accompanying notable decrement in precision. Systematic ablation analysis unequivocally confirmed the individual contribution of each architectural component, demonstrating that ASPP, the R2 blocks, dual attention mechanisms, and Tversky loss each address a distinct challenge inherent in DWI-based infarct segmentation. The Wilcoxon signed-rank test, paired t-test, McNemar test, and bootstrap confidence intervals also confirmed the reliability of the obtained results. Although the Cohen's d values were observed to be low, this outcome primarily correlates with the already high performance of the comparator models. These findings collectively suggest that EPRA U-Net may provide a robust and highly accurate foundation for DWI-based infarct segmentation within clinical practice.

**Acknowledgements**

**Funding:** This research received financial support from the Scientific and Technological Research Council of Türkiye (TÜBİTAK) 1002-A Short-Term Support Module (Grant number: 124E595).

**Conflicts of interest statement:** None of the four authors have any conflicts of interest to disclose.

**Data availability statement:** The datasets generated and/or analysed during the current study are available in the [Github] repository, [https://github.com/ulutashasan/Enfarkt_EpraUnet]

**Ethics declarations:** The dataset used in this study was retrospectively collected from the Department of Radiology at Yozgat Bozok University Training and Research Hospital. Ethical approval for this study was obtained by the Clinical Research Ethics Committee of Yozgat Bozok University (Protocol No:36/27).

**Consent to Participate:** Informed consent was waived by the Ethics Committee of Yozgat Bozok University due to the retrospective nature of the study and the use of fully anonymized data. The study was conducted in accordance with the ethical standards of the institutional research committee and with the 1964 Helsinki Declaration and its later amendments.

**Author Contributions**

**Hasan Ulutas:** Supervision, Software, methodology, manuscript review and editing.

**M. Emin Sahin:** Conceptualization, methodology, supervision, manuscript writing and revision.

**Esra Yuce:** Dataset curation, investigation.

**M. Fatih Erkoc:** Dataset curation, writing original draft, resources

**Turker Tuncer:** manuscript writing and revision, resources

**Sengul Dogan:** manuscript writing and revision, investigation

**Serkan Kiranyaz**: Software, supervision

## References

[1] E. Roldán-Valadéz and M. López-Mejía, "Current concepts on magnetic resonance imaging (MRI) perfusion-diffusion assessment in acute ischaemic stroke: a review & an update for the clinicians.," Indian J Med Res. , vol. 140, no. 6, pp. 717–28, Dec. 2014, PMID: 25758570; PMCID: PMC4365345.
[2] V. L. Feigin et al., "World Stroke Organization: Global Stroke Fact Sheet 2025," International Journal of Stroke, vol. 20, no. 2, pp. 132–144, Dec. 2024, doi: 10.1177/17474930241308142.
[3] A. Elhanashi, P. Dini, S. Saponara, and Q. Zheng, "TeleStroke: real-time stroke detection with federated learning and YOLOv8 on edge devices," Journal of Real-Time Image Processing, vol. 21, no. 4, Jun. 2024, doi: 10.1007/s11554-024-01500-1.
[4] N. Nagaraja, "Diffusion weighted imaging in acute ischemic stroke: A review of its interpretation pitfalls and advanced diffusion imaging application," Journal of the Neurological Sciences , vol. 425, pp. 117435–117435, Apr. 2021, doi: 10.1016/j.jns.2021.117435.
[5] V. T. Heeralal, S. E. Chadee, B. Ilyaev, R. Ilyaev, and S. Ilyayeva, "Artificial Intelligence in Stroke Care: A Narrative Review of Diagnostic, Predictive, and Workflow Applications," Cureus , vol. 17, no. 9, Sep. 2025, doi: 10.7759/cureus.93430.
[6] O. Ronneberger, P. Fischer, and T. Brox, "U-Net: Convolutional Networks for Biomedical Image Segmentation," in Lecture notes in computer science, Springer Science+Business Media, 2015, pp. 234–241. doi: 10.1007/978-3-319-24574-4_28.
[7] Z. Zhou, M. M. R. Siddiquee, N. Tajbakhsh, and J. Liang, "UNet++: A Nested U-Net Architecture for Medical Image Segmentation," Lecture notes in computer science , vol. 11045, pp. 3–11, Jan. 2018, doi: 10.1007/978-3-030-00889-5_1.
[8] O. Oktay et al., "Attention U-Net: Learning Where to Look for the Pancreas," arXiv (Cornell University), Apr. 2018, doi: 10.48550/arxiv.1804.03999.
[9] M. Z. Alom, Md. M. Hasan, C. Yakopcic, T. M. Taha, and V. K. Asari, "Recurrent Residual Convolutional Neural Network based on U-Net (R2U-Net) for Medical Image Segmentation," arXiv (Cornell University), Feb. 2018, doi: 10.48550/arxiv.1802.06955.
[10] J. Chen et al., "TransUNet: Transformers Make Strong Encoders for Medical Image Segmentation," arXiv (Cornell University), Feb. 2021, doi: 10.48550/arxiv.2102.04306.
[11] H. Cao et al., "Swin-Unet: Unet-like Pure Transformer for Medical Image Segmentation," arXiv (Cornell University), May 2021, doi: 10.48550/arxiv.2105.05537.
[12] R. Azad et al., "Medical Image Segmentation Review: The Success of U-Net," IEEE Transactions on Pattern Analysis and Machine Intelligence, vol. 46, no. 12. IEEE Computer Society, pp. 10076–10095, Aug. 21, 2024. doi: 10.1109/tpami.2024.3435571.
[13] Ö. Çiçek, A. Abdulkadir, S. S. Lienkamp, T. Brox, and O. Ronneberger, "3D U-Net: Learning Dense Volumetric Segmentation from Sparse Annotation," in Lecture notes in computer science, Springer Science+Business Media, 2016, pp. 424–432. doi: 10.1007/978-3-319-46723-8_49.
[14] F. Milletarì, N. Navab, and S. Ahmadi, "V-Net: Fully Convolutional Neural Networks for Volumetric Medical Image Segmentation," pp. 565–571, Oct. 2016, doi: 10.1109/3dv.2016.79.
[15] K. Kamnitsas et al., "Efficient multi-scale 3D CNN with fully connected CRF for accurate brain lesion segmentation," Medical Image Analysis, vol. 36, pp. 61–78, Oct. 2016, doi: 10.1016/j.media.2016.10.004.

[16] F. Isensee, P. F. Jaeger, S. A. A. Kohl, J. Petersen, and K. H. Maier-Hein, “nnU-Net: a self-configuring method for deep learning-based biomedical image segmentation,” Nature Methods, vol. 18, no. 2, pp. 203–211, Dec. 2020, doi: 10.1038/s41592-020-01008-z.
[17] L.-C. Chen, Y. Zhu, G. Papandreou, F. Schroff, and H. Adam, “Encoder-Decoder with Atrous Separable Convolution for Semantic Image Segmentation,” in Lecture notes in computer science, Springer Science+Business Media, 2018, pp. 833–851. doi: 10.1007/978-3-030-01234-2_49.
[18] H. Zhao, J. Shi, X. Qi, X. Wang, and J. Jia, “Pyramid Scene Parsing Network,” pp. 6230–6239, Jul. 2017, doi: 10.1109/cvpr.2017.660.
[19] A. Vaswani et al., “Attention is All you Need,” in Adv. Neural Inf. Process. Syst., 2017.
[20] S. Woo, J. Park, J. Lee, and I. S. Kweon, “CBAM: Convolutional Block Attention Module,” in Lecture notes in computer science, Springer Science+Business Media, 2018, pp. 3–19. doi: 10.1007/978-3-030-01234-2_1.
[21] J. Fu et al., “Dual Attention Network for Scene Segmentation,” pp. 3141–3149, Jun. 2019, doi: 10.1109/cvpr.2019.00326.
[22] C. H. Sudre, W. Li, T. Vercauteren, S. Ourselin, and M. J. Cardoso, “Generalised Dice Overlap as a Deep Learning Loss Function for Highly Unbalanced Segmentations,” Lecture notes in computer science, vol. 2017, pp. 240–248, Jan. 2017, doi: 10.1007/978-3-319-67558-9_28.
[23] E. Tappeiner, M. Welk, and R. Schubert, “Tackling the class imbalance problem of deep learning-based head and neck organ segmentation,” International Journal of Computer Assisted Radiology and Surgery, vol. 17, no. 11, pp. 2103–2111, May 2022, doi: 10.1007/s11548-022-02649-5.
[24] S. S. M. Salehi, D. Erdoğmuş, and A. Gholipour, “Tversky Loss Function for Image Segmentation Using 3D Fully Convolutional Deep Networks,” Lecture notes in computer science, pp. 379–387, Jan. 2017, doi: 10.1007/978-3-319-67389-9_44.
[25] J. Terven, D. Córdova-Esparza, J.-A. Romero-González, A. Ramírez-Pedraza, and E. A. Chávez-Urbiola, “A comprehensive survey of loss functions and metrics in deep learning,” Artificial Intelligence Review, vol. 58, no. 7, Apr. 2025, doi: 10.1007/s10462-025-11198-7.
[26] N. Abraham and N. Khan, “A Novel Focal Tversky Loss Function With Improved Attention U-Net for Lesion Segmentation,” pp. 683–687, Apr. 2019, doi: 10.1109/isbi.2019.8759329.
[27] B. Menze et al., “The Multimodal Brain Tumor Image Segmentation Benchmark (BRATS),” IEEE Transactions on Medical Imaging, vol. 34, no. 10, pp. 1993–2024, Dec. 2014, doi: 10.1109/tmi.2014.2377694.
[28] A. Rahman et al., “Deep Learning-Driven Segmentation of Ischemic Stroke Lesions Using Multi-Channel MRI,” arXiv (Cornell University), Jan. 2025, doi: 10.48550/arxiv.2501.02287.
[29] J. Åkesson, J. Töger, and E. Heiberg, “Random effects during training: Implications for deep learning-based medical image segmentation,” Computers in Biology and Medicine, vol. 180, pp. 108944–108944, Aug. 2024, doi: 10.1016/j.compbiomed.2024.108944.
[30] D. Müller, I. Soto-Rey, and F. Krämer, “Robust chest CT image segmentation of COVID-19 lung infection based on limited data,” Informatics in Medicine Unlocked, vol. 25, pp. 100681–100681, Jan. 2021, doi: 10.1016/j.imu.2021.100681.
[31] S. Patil, R. Rossi, D. Jabrah, and K. Doyle, “Detection, Diagnosis and Treatment of Acute Ischemic Stroke: Current and Future Perspectives,” Frontiers in Medical Technology, vol. 4. Frontiers Media, Jun. 24, 2022. doi: 10.3389/fmedt.2022.748949.
[32] P. Skalski, "Make Sense: Free to use online annotation tool," GitHub, 2019. [Online]. Available: https://github.com/SkalskiP/make-sense.
[33] E. Yağış et al. , “Effect of data leakage in brain MRI classification using 2D convolutional neural networks,” Scientific Reports, vol. 11, no. 1, Nov. 2021, doi: 10.1038/s41598-021-01681-w.
[34] A. Kumar et al., “A Flexible 2.5D Medical Image Segmentation Approach with In-Slice and Cross-Slice Attention,” arXiv (Cornell University), Apr. 2024, doi: 10.48550/arxiv.2405.00130.
[35] M. Tan and Q. V. Le, “EfficientNet: Rethinking Model Scaling for Convolutional Neural Networks,” arXiv (Cornell University), May 2019, doi: 10.48550/arxiv.1905.11946.
[36] K. He, X. Zhang, S. Ren, and J. Sun, “Deep Residual Learning for Image Recognition,” pp. 770–778, Jun. 2016, doi: 10.1109/cvpr.2016.90.
[37] L.-C. Chen, G. Papandreou, I. Kokkinos, K. Murphy, and A. Yuille, “DeepLab: Semantic Image Segmentation with Deep Convolutional Nets, Atrous Convolution, and Fully Connected CRFs,” IEEE Transactions on Pattern Analysis and Machine Intelligence, vol. 40, no. 4, pp. 834–848, Apr. 2017, doi: 10.1109/tpami.2017.2699184.
[38] K. Dutta, “Densely Connected Recurrent Residual (Dense R2UNet) Convolutional Neural Network for Segmentation of Lung CT Images,” arXiv (Cornell University) , Feb. 2021, doi: 10.48550/arxiv.2102.00663.
[39] L. Burrows, J. Patel, A. I. Islim, M. D. Jenkinson, S. J. Mills, and K. Chen, “A semi-automatic segmentation method for meningioma developed using a variational approach model,” The Neuroradiology Journal, vol. 37, no. 2, pp. 199–205, Dec. 2023, doi: 10.1177/19714009231224442.
[40] M. L. Seghier, “Image Segmentation Evaluation With the Dice Index: Methodological Issues,” International Journal of Imaging Systems and Technology, vol. 34, no. 6, Oct. 2024, doi: 10.1002/ima.23203.

[41] D. Müller, I. Soto-Rey, and F. Krämer, “Towards a guideline for evaluation metrics in medical image segmentation,” BMC Research Notes, vol. 15, no. 1. BioMed Central, Jun. 20, 2022. doi: 10.1186/s13104-022-06096-y.
[42] A. Reinke et al, “Common Limitations of Image Processing Metrics: A Picture Story,” arXiv (Cornell University), vol. 2021, Apr. 2021, doi: 10.48550/arxiv.2104.05642.
[43] M. E. Şahin, H. Ulutaş, E. Yüce, and M. F. Erkoç, “Detection and classification of COVID-19 by using faster R-CNN and mask R-CNN on CT images,” Neural Computing and Applications, vol. 35, no. 18, pp. 13597–13611, Mar. 2023, doi: 10.1007/s00521-023-08450-y.
[44] N. Siddique, S. Paheding, C. Elkin, and V. Devabhaktuni, “U-Net and Its Variants for Medical Image Segmentation: A Review of Theory and Applications,” IEEE Access, vol. 9, pp. 82031–82057, Jan. 2021, doi: 10.1109/access.2021.3086020.
[45] M. E. Şahin, “Deep learning-based approach for detecting COVID-19 in chest X-rays,” Biomedical Signal Processing and Control, vol. 78, pp. 103977–103977, Jul. 2022, doi: 10.1016/j.bspc.2022.103977.
[46] M. Yeung, L. Rundo, N. Yang, E. Sala, C. Schönlieb, and G. Yang, “Calibrating the Dice Loss to Handle Neural Network Overconfidence for Biomedical Image Segmentation,” Journal of Digital Imaging, vol. 36, no. 2, pp. 739–752, Dec. 2022, doi: 10.1007/s10278-022-00735-3.
[47] S. Duraivenkatesh, A. Narayan, V. Srikanth, and A. F. Made, “Retinoblastoma Detection via Image Processing and Interpretable Artificial Intelligence Techniques,” medRxiv (Cold Spring Harbor Laboratory), May 2023, doi: 10.1101/2023.05.02.23289419.
[48] W. Zhang and S. Ray, “From coarse to fine: a deep 3D probability volume contours framework for tumour segmentation and dose painting in PET images,” Frontiers in Radiology, vol. 3, Sep. 2023, doi: 10.3389/fradi.2023.1225215.
[49] F. Wilcoxon, “Individual Comparisons by Ranking Methods,” Biometrics Bulletin, vol. 1, no. 6, pp. 80–80, Dec. 1945, doi: 10.2307/3001968.
[50] A. J. Vickers, "Parametric versus non-parametric statistics in the analysis of randomized trials with non-normally distributed data," BMC Medical Research Methodology, vol. 5, no. 1, p. 35, 2005, doi: 10.1186/1471-2288-5-35.
[51] Q. McNemar, “Note on the Sampling Error of the Difference Between Correlated Proportions or Percentages,” Psychometrika, vol. 12, no. 2, pp. 153–157, Jun. 1947, doi: 10.1007/bf02295996.
[52] B. Efron and R. Tibshirani, An Introduction to the Bootstrap. 1994. doi: 10.1201/9780429246593.
[53] J. Cohen, Statistical Power Analysis for the Behavioral Sciences, 2nd ed. Hillsdale, NJ: Lawrence Erlbaum Associates, 1988.
[54] C. A. Beam, “Resolving power: a general approach to compare the distinguishing ability of threshold-free evaluation metrics,” Machine Learning, vol. 114, no. 1, Jan. 2025, doi: 10.1007/s10994-024-06723-8.
[55] M. Haimerl and C. Reich, “Risk-based Evaluation of ML Classification Methods Used for Medical Devices,” Research Square (Research Square), Sep. 2023, doi: 10.21203/rs.3.rs-3317894/v1.
[56] P. M. Konrad, A.-A. Popa, Y. Sabzehmeidani, L. Zhong, E. A. Liehn, and S. Ayvaz, “Challenges in Deep Learning-Based Small Organ Segmentation: A Benchmarking Perspective for Medical Research with Limited Datasets,” arXiv (Cornell University), Sep. 2025, doi: 10.48550/arxiv.2509.05892.
[57] J. W. Hoving et al., “Accuracy of CT perfusion ischemic core volume and location estimation: A comparison between four ischemic core estimation approaches using syngo.via,” PLoS ONE, vol. 17, no. 8, Aug. 2022, doi: 10.1371/journal.pone.0272276.
[58] J. H. Han et al., “Automated identification of thrombectomy amenable vessel occlusion on computed tomography angiography using deep learning,” Frontiers in Neurology, vol. 15, Jul. 2024, doi: 10.3389/fneur.2024.1442025.
[59] A. Tsuchida, P. Boutinaud, V. Verrecchia, C. Tzourio, S. Debette, and M. Joliot, “Early detection of white matter hyperintensities using SHIVA-WMH detector,” Human Brain Mapping, vol. 45, no. 1, Dec. 2023, doi: 10.1002/hbm.26548.
[60] P. V. Nayantara, S. Kamath, R. Kadavigere, and K. Manjunath, “Automatic Liver Segmentation from Multiphase CT Using Modified SegNet and ASPP Module,” SN Computer Science, vol. 5, no. 4, Mar. 2024, doi: 10.1007/s42979-024-02719-2.
[61] Sonaglioni, A., Di Cara, M., Nicolosi, G. L., Eusebio, A., Bordonali, M., Santalucia, P., & Lombardo, M. "Rapid risk stratification of acute ischemic stroke patients in the emergency department: the incremental prognostic role of left atrial reservoir strain." *Journal of Stroke and Cerebrovascular Diseases* 30.11 (2021): 106100.
[62] N. Wu, D. Jia, Z. Li, and Z. He, “Weak Edge Target Segmentation Network Based on Dual Attention Mechanism,” Applied Sciences, vol. 14, no. 19, pp. 8963–8963, Oct. 2024, doi: 10.3390/app14198963.